\pdfoutput=1
\documentclass{article}

\usepackage{iclr2027_conference,times}

\iclrfinalcopy

\usepackage[T1]{fontenc}
\usepackage[utf8]{inputenc}

\usepackage{graphicx}
\usepackage{microtype}
\usepackage{inconsolata}
\usepackage{courier}

\usepackage{amsmath}
\usepackage{amsfonts}
\usepackage{amssymb}
\usepackage{mathtools}
\usepackage{latexsym}
\usepackage{wrapfig}
\usepackage{bbm}
\usepackage{tabularx}
\usepackage{array}

\usepackage{booktabs}
\usepackage{multirow}
\usepackage{makecell}
\usepackage[table]{xcolor}

\definecolor{Gray}{gray}{0.9}
\definecolor{tintred}{HTML}{F6D7D9}
\definecolor{tintgray}{HTML}{E8E8E8}
\definecolor{tintblue}{HTML}{DAE8FC}
\definecolor{darkblue}{RGB}{0,0,180}
\definecolor{darkred}{RGB}{180,0,0}

\usepackage{algorithm}
\usepackage{algpseudocode}

\usepackage{enumitem}
\usepackage{pifont}
\usepackage{soul}
\usepackage{xspace}
\usepackage{caption}
\usepackage{marvosym}

\usepackage{url}
\usepackage{hyperref}

\newcolumntype{Y}{>{\raggedright\arraybackslash}X}
\newcommand{\stagearrow}{\,\(\rightarrow\)\,}

\makeatletter
\DeclareRobustCommand\onedot{\futurelet\@let@token\@onedot}
\def\@onedot{\ifx\@let@token.\else.\null\fi\xspace}
\def\eg{\emph{e.g}\onedot}

 \def\vs{\emph{vs}\onedot}

\makeatother
\newcommand{\name}{\textsc{StructRL}}
\newcommand{\codeurl}{https://github.com/amazon-science/StructRL}

\title{
{\name}: Online Structured Reinforcement Learning for
Long-Horizon Vision-Language-Action Tasks
}

\author{
\makebox[\textwidth][c]{%
Ziyi Yin\textsuperscript{$\spadesuit$$\dagger$$*$}
\quad
Sangmin Woo\textsuperscript{$\heartsuit$$\dagger$}
\quad
Kang Zhou\textsuperscript{$\heartsuit$\Letter}
\quad
Sungyeon Kim\textsuperscript{$\heartsuit$}
\quad
Aosong Feng\textsuperscript{$\heartsuit$}
} \\
\makebox[\textwidth][c]{%
\textbf{Haibo Ding}\textsuperscript{$\heartsuit$}
\quad
\textbf{Luke Huan}\textsuperscript{$\heartsuit$}
} \\
\makebox[\textwidth][c]{%
\textsuperscript{$\spadesuit$}The Pennsylvania State University
\qquad
\textsuperscript{$\heartsuit$}Amazon AWS AI
} \\
\makebox[\textwidth][c]{%
\texttt{ziyiyin@psu.edu}
\quad
\texttt{\{sangminw, zhoukang\}@amazon.com}
}
}

\begin{document}

\maketitle

\lhead{}
\renewcommand{\headrulewidth}{0pt}

\begingroup
\renewcommand{\thefootnote}{$\dagger$}\footnotetext{Co-first authors.}
\renewcommand{\thefootnote}{$*$}\footnotetext{Work done during an internship at Amazon.}
\endgroup

\addtocontents{toc}{\protect\setcounter{tocdepth}{-1}}   %
\begin{abstract}

Vision-language-action (VLA) models perform well on shorter-horizon manipulation tasks but still struggle with long-horizon tasks that require multiple dependent manipulations from a single command.
Online reinforcement learning (RL) can improve these policies through environment interaction, yet many existing methods provide reward only after the complete task succeeds.
However, such terminal supervision is sparse and does not distinguish early failures from rollouts that make substantial partial progress.
We propose {\name}, an online RL framework that constructs structured intermediate supervision from verifiable subtask completions.
{\name} decomposes each task into verifiable subtasks, grants intermediate rewards only after the prerequisite subtasks have been completed, and scales each reward according to completion pace.
Across RoboCasa365 and LIBERO-Long with GR00T-N1.5 and $\pi_{0.5}$, {\name} consistently outperforms evaluated online RL baselines.
These results show that verifiable, structured intermediate rewards improve long-horizon VLA post-training.
Code is available at \url{\codeurl}.

\end{abstract}

\begin{figure}[h]
    \centering
    \vspace{-2mm}
    \includegraphics[width=\linewidth]{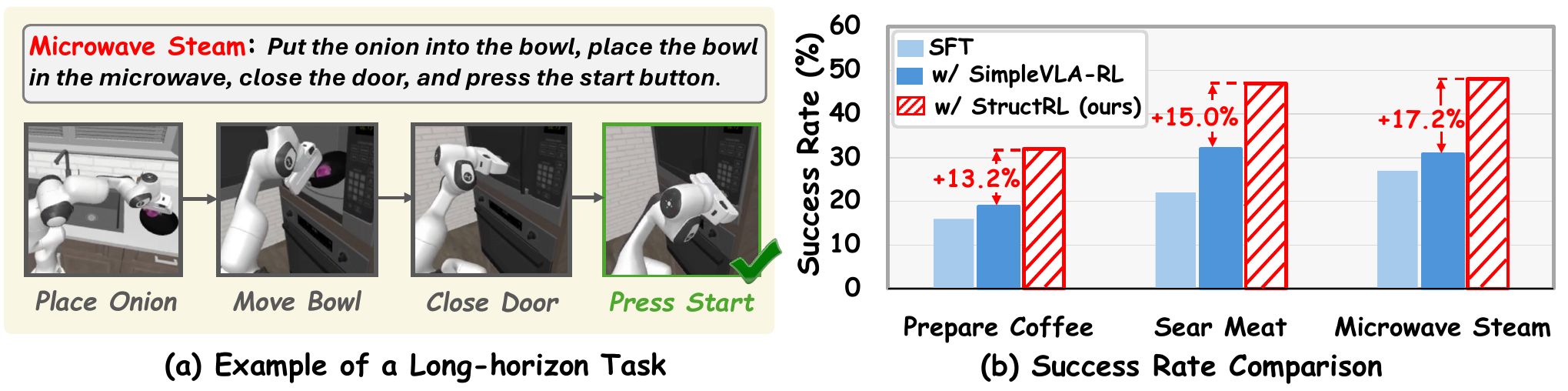}
    \vspace{-5mm}
    \caption{(a) An example long-horizon VLA task from RoboCasa365~\citep{robocasa365}. (b) Success rates (\%) of GR00T-N1.5 on three representative composite tasks. {\name} consistently outperforms both the SFT policy and the RL baseline SimpleVLA-RL~\citep{li2026simplevla}.
    }
    \label{fig:teaser}
\end{figure}

\section{Introduction}

Vision-language-action (VLA) models map visual observations and language instructions to low-level robot actions~\citep{black2025pi0,black2025pi05,bjorck2025gr00t,kim2026rldx}. Recent models have shown strong capabilities in household manipulation~\citep{black2025pi05}, humanoid control~\citep{bjorck2025gr00t}, and dexterous manipulation~\citep{geminirobotics2025}.
Despite this progress, VLA evaluation still focuses primarily on tasks such as picking up an object and placing it at a target location~\citep{liu2023libero,li2024simpler}. These instructions typically specify a single self-contained skill rather than an extended sequence of dependent actions~\citep{han2025robocerebra,mees2022calvin}.

A more capable VLA should execute an extended task from a single instruction, without requiring a new command after each manipulation. We refer to these problems as \emph{long-horizon tasks}. As illustrated in Figure~\ref{fig:teaser}, they require the policy to compose multiple dependent skills, such as grasping several objects, navigating to a target location, and operating articulated fixtures in the correct order. Long-horizon VLA policies are commonly trained through supervised fine-tuning (SFT) on human demonstrations~\citep{robocasa365}, yet their task success rates remain limited. Because SFT exposes the policy primarily to demonstrated states, small errors during a long rollout can move it into poorly covered states from which recovery is difficult.

Online reinforcement learning (RL) offers a natural way to address this distribution shift by allowing the policy to train on states encountered during its own interactions~\citep{zang2026rlinfvla,robocasa365}. Most online VLA RL methods, however, rely on a terminal reward that is issued only when the entire task succeeds~\citep{robocasa365,wang2026policytrim,li2026simplevla}. This signal becomes increasingly sparse as the task horizon grows and successful rollouts become rare. More importantly, it does not represent partial progress: a rollout that completes every step except the last receives the same return as one that fails near the beginning. Long-horizon VLA training therefore requires a denser signal that can identify meaningful progress before final success.

Learned reward models provide denser supervision by estimating intermediate task progress~\citep{shu2025rftf,zhang2026vlac,tan2025robodopamine,liang2026robometer}. A scalar progress estimate, however, does not by itself specify which prerequisite events make a detected completion valid. In long-horizon manipulation, an event may appear locally useful but fail to advance the task because a required prerequisite has not yet been completed.
Therefore, dense supervision should encode both whether an event occurred and whether it constitutes valid progress toward the final goal given the events completed so far.

To provide dense supervision while respecting the dependency structure of long-horizon tasks, we propose {\name}, an online RL framework that constructs intermediate rewards. {\name} applies an automatic pipeline to decompose each instruction into a set of subtasks whose completion can be directly verified from the environment state, then organizes them into ordered dependency groups. This structure matters because satisfying a local completion does not always constitute valid task progress. For example, closing a box before the required objects have been placed inside should not receive intermediate credit.

{\name} rewards each verified subtask completion and shapes that reward with two mechanisms.
\emph{Structure-aware reward gating} determines whether a detected completion is eligible for intermediate credit: a subtask is rewarded only after all of its prerequisites have been satisfied. \emph{Dynamic reward pacing} determines the magnitude of that credit, assigning larger rewards to subtasks completed more quickly relative to the demonstrations. The resulting chunk-level rewards are used to optimize the VLA with Proximal Policy Optimization (PPO)~\citep{schulman2017ppo}. Together, these mechanisms provide dense supervision while keeping reward assignment verifiable and consistent with task dependencies.

We evaluate {\name} on RoboCasa365 and LIBERO-Long using both GR00T-N1.5 and $\pi_{0.5}$. Across both benchmarks and backbones, {\name} consistently outperforms the evaluated online RL baselines. On RoboCasa365 with GR00T-N1.5, for example, {\name} reaches 49.1\% success rate (SR), compared with 38.6\% for SFT and 41.5\% for the strongest evaluated online RL baseline. The component ablation shows that verified subtask rewards provide most of this gain, with gating and pacing adding further improvements. Additional results show that the structured reward is compatible with GRPO and remains effective on shorter-horizon LIBERO suites. Collectively, these results support structured intermediate supervision as an effective approach to long-horizon VLA post-training.

\begin{figure*}[t]
\centering
\includegraphics[width=1.0\textwidth]{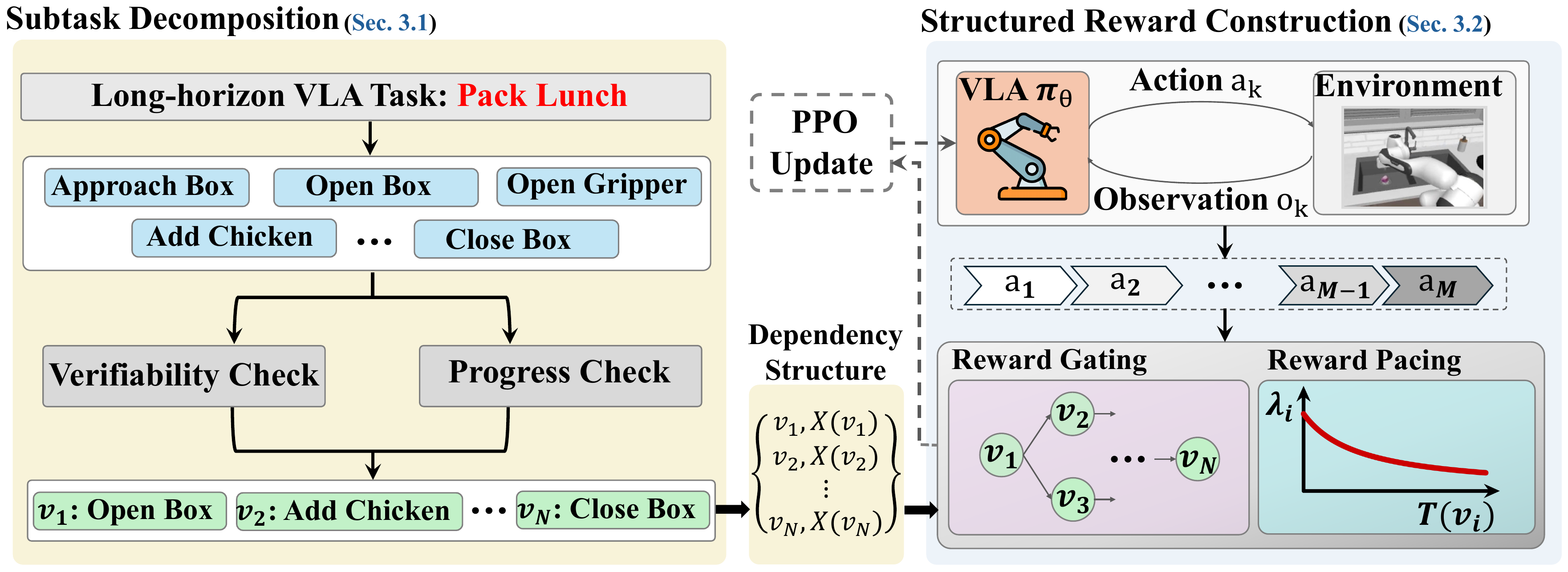}
\caption{Overview of {\name}.
\textbf{Left:} An LLM decomposes the task command into candidate subtasks and assigns a dependency structure to the retained subtasks, represented by prerequisite sets $X(v_i)$. The verifiability check removes candidates without a reliable binary completion criterion (\eg, \emph{Approach Box}), while the progress check removes candidates that do not by themselves indicate progress toward task completion (\eg, \emph{Open Gripper}). \textbf{Right:} Structure-aware reward gating uses this dependency structure to determine whether a detected completion is eligible for reward, while dynamic reward pacing determines the magnitude of that reward. The resulting chunk-level rewards are used to optimize the VLA with PPO. The training pseudocode is shown in Appendix~\ref{app:training}.
}
\label{fig:framework}
\vspace{-2mm}
\end{figure*}

\section{Preliminaries}\label{sec:prelim}
In this section, we formalize action-chunk VLA interaction and PPO training, establishing the notation used by the structured reward formulation in Section~\ref{sec:method}.

\noindent\textbf{VLA Rollouts.}
We model VLA interaction with the environment at the action-chunk level. At chunk index $k$, the policy $\pi_\theta$ receives an observation $o_k$ containing multi-view images, the language instruction, and proprioceptive state, and uses its action expert to generate an action chunk $a_k\in\mathbb{R}^{L\times d}$. Here, $L$ is the number of consecutive low-level actions in the chunk, and $d$ is the number of controllable robot degrees of freedom.\footnote{We use flow-matching VLAs (\eg, GR00T-N1.5 and $\pi_{0.5}$) as the running examples; the same action-chunk formulation applies to autoregressive VLAs, whose token-level likelihoods are directly available.} The environment executes the complete chunk before returning $o_{k+1}$. The resulting rollout is $\tau=(o_1,a_1,\ldots,o_M,a_M)$, containing $M$ action chunks, which is used for RL training.

\noindent\textbf{PPO Training.}
We next explain how the collected rollouts $\tau$ are used to optimize
$\pi_\theta$, taking PPO~\citep{schulman2017ppo} as an example. For VLA training, PPO
operates at the chunk level: for each chunk $a_k$ in $\tau$, it maximizes
the clipped surrogate objective

{\small
\begin{equation}
\mathcal{L}(\theta)=\mathbb{E}_k\Big[\min\big(\rho_k A_k,\;
\operatorname{clip}(\rho_k,1{-}\epsilon,1{+}\epsilon)\,A_k\big)\Big],
\label{eq:ppo}
\end{equation}}%
where $\epsilon$ is the standard PPO clipping threshold,
$\rho_k=\pi_\theta(a_k\mid o_k)/\pi_{\theta_{\mathrm{old}}}(a_k\mid o_k)$
is the importance sampling ratio, and $A_k$ is the advantage estimate for
each chunk $a_k$, computed from the chunk-level reward sequence using a learned
critic. Under the standard terminal-only reward setting, intermediate chunks
receive zero reward, while the final chunk receives a positive reward only
upon successful task completion.

For flow-matching VLAs, computing $\rho_k$ is nontrivial because deterministic
ODE sampling does not directly provide the chunk likelihood
$\pi_\theta(a_k\mid o_k)$. Following prior work~\citep{zang2026rlinfvla}, we
inject Gaussian noise into each denoising step to convert the ODE sampling
process into an SDE, making the chunk likelihood tractable. This enables the
likelihood ratio required by PPO to be evaluated. We next introduce {\name},
which replaces the terminal-only reward with structured intermediate rewards
while retaining the PPO objective in Eq.~\eqref{eq:ppo}.

\section{Methodology}\label{sec:method}

Let $c$ denote a long-horizon task command and $\pi_\theta$ the VLA policy to be fine-tuned through online RL. As illustrated in Figure~\ref{fig:framework}, {\name} operates in two stages. First, an LLM decomposes $c$ into a set of verifiable, goal-aligned subtasks and organizes them into ordered dependency groups.
Second, {\name} converts this structure into chunk-level rewards through structure-aware reward gating and dynamic reward pacing, then optimizes $\pi_\theta$ with PPO (Section~\ref{sec:reward}).

\subsection{Subtask Decomposition}\label{sec:decomp}
\noindent{\textbf{Semantic decomposition}}. A long-horizon command may combine heterogeneous skills, including grasping, base navigation, and articulated-object manipulation. We decompose the command into intermediate subtasks that satisfy two requirements: completion can be detected using a binary environment-state criterion, and the completed subtask represents progress toward the final task goal.

For each command $c$, we prompt an LLM to propose natural-language candidate subtasks, such as \emph{open the box}, and to return only candidates that pass two checks:
(i) \textit{Verifiability Check}: completion can be detected from the environment state using a binary criterion.
(ii) \textit{Progress Check}: satisfying that criterion represents progress toward the final task goal rather than an incidental or non-progressive behavior.

Figure~\ref{fig:framework} illustrates both checks for the \emph{Pack Lunch} task. \emph{Approach Box} fails the Verifiability Check because its completion is difficult to define using a reliable binary environment-state criterion, while \emph{Open Gripper} fails the Progress Check because it does not by itself indicate a successful task progress and may also occur during a failed attempt. The retained subtasks form $\mathcal{V}=\{v_1,\ldots,v_N\}$, where $N$ is the number of subtasks.

\noindent{\textbf{Dependency structure assignment}}.
A flat set of subtasks is insufficient to fully characterize progress in a
long-horizon task, because the subtasks in $\mathcal{V}$ are not mutually
independent: some must be completed before others, while others may be executed
in arbitrary order. Ignoring these relations can treat an out-of-order
completion as valid task progress. For example, the \emph{Pack Lunch}
decomposition follows the dependency pattern
$\emph{Open Box}\rightarrow\{\emph{Add Chicken},~\emph{Add Apple}\}
\rightarrow\emph{Close Box}$, where the two additions may occur in either
order, but \emph{Close Box} should only be considered valid progress after both
additions have been completed.

Accordingly, the final LLM output organizes the retained subtasks into ordered
dependency groups.
Subtasks within the same group may be completed in any
order, whereas all subtasks in an earlier group must be completed before those
in a later group. We convert this stage-wise ordering into prerequisite sets
for reward construction. For each subtask $v_i\in\mathcal{V}$, we define
$X(v_i)\subseteq\mathcal{V}$ as the set of subtasks that must be completed
before $v_i$, with $X(v_i)=\varnothing$ for subtasks without prerequisites:
\begin{equation}
\mathcal{V}
\;\longrightarrow\;
\{(v_1,X(v_1)),\dots,(v_N,X(v_N))\}.
\label{eq:decomp}
\end{equation}
This representation determines when each detected subtask completion becomes
eligible for intermediate reward. Appendix~\ref{app:decomp_prompt} provides
the complete prompt and generated decompositions, and
Section~\ref{exp:decomposition} analyzes decomposition granularity. We next
describe how this structured decomposition is converted into rewards.

\subsection{Structured Reward Construction}\label{sec:reward}
Based on the decomposed subtasks and the dependency structure among
them, {\name} constructs the reward for each action chunk through two decisions. Structure-aware reward gating determines whether a detected subtask completion is eligible for reward, based on the task dependencies. Dynamic reward pacing determines the magnitude of each eligible reward, based on completion pace. We describe these eligibility and magnitude components in turn.

\subsubsection{Structure-Aware Reward Gating}\label{sec:gating}

Consider a rollout $\tau$ with action chunks $\{a_1,\ldots,a_M\}$. A detected completion of $v_i$ is eligible for intermediate reward only if every prerequisite in $X(v_i)$ has already been completed and $v_i$ has not previously been rewarded. Thus, each subtask can contribute intermediate reward at most once. The reward assigned to each chunk $a_k$ is therefore defined by
\begin{equation}
r_k =
\underbrace{\textstyle\sum_{i=1}^{N}
\mathbbm{1}_k(v_i, X(v_i)) \cdot \lambda_i}
_{\text{\footnotesize subtask rewards}}
+
\underbrace{\mathbbm{1}_k(c) \cdot \lambda_c
\vphantom{\textstyle\sum_{i=1}^{N}}}
_{\mathclap{\text{\footnotesize terminal reward}}}.
\label{eq:step-reward}
\end{equation}
Here, $\mathbbm{1}_k(v_i,X(v_i))=1$ when completion of $v_i$ is detected at chunk $a_k$, every prerequisite in $X(v_i)$ has been completed, and $v_i$ has not previously been rewarded.
The terminal indicator $\mathbbm{1}_k(c)=1$ when the complete task first succeeds at chunk $a_k$. We use a fixed terminal reward $\lambda_c$ and determine each intermediate reward $\lambda_i$ dynamically from its completion pace.

\subsubsection{Dynamic Reward Pacing}\label{sec:pacing}

A straightforward choice of $\lambda_i$ is to use a fixed reward for
all subtasks. However, such a design does not explicitly distinguish fast,
direct completions from delayed ones involving unnecessary wandering, which
can blur credit assignment for task progress in long-horizon rollouts. Therefore, {\name} instead scales the reward according to completion pace:
\begin{equation}
\lambda_i
=\beta\cdot\frac{1}{1+T(v_i)/T_d(v_i)}.
\label{eq:pace}
\end{equation}
Here, $T(v_i)$ is the number of action chunks elapsed since the previous rewarded subtask completion; for the first rewarded subtask, it is measured from the beginning of the rollout.
For each subtask $v_i$, we compute $T_d(v_i)$ from the demonstration interval that begins when all prerequisites of $v_i$ have first become complete and ends when $v_i$ is completed. We average this interval across demonstrations in which both events are observed.
The scale parameter $\beta$ upper-bounds the reward for one subtask.

Finally, we train the VLA with the PPO objective introduced in
Eq.~\eqref{eq:ppo}, resulting in the optimized policy $\pi_\theta$.

\subsubsection{Extension to GRPO}\label{sec:method_grpo}
PPO is our default optimizer, but the same structured rewards can be used with Group Relative Policy Optimization (GRPO)~\citep{shao2024deepseekmath}.
For each task instruction paired with one fixed initial state, we collect a group of $G=8$ rollouts. We run $256$ environments in parallel for two rollout epochs, so each training iteration collects $64$ groups, or $512$ rollouts, the same number as PPO.
Within each rollout $\tau$, we sum the chunk-level rewards $R(\tau)=\sum_{k=1}^{M}r_k$, and GRPO normalizes these returns within each rollout group to compute group-relative advantages.
Section~\ref{sec:extend_grpo} reports the corresponding results.

\section{Experiments}\label{sec:exp}

\subsection{Experimental Setup}\label{sec:exp_setup}
\noindent\textbf{Benchmarks.}
We evaluate {\name} on two manipulation benchmarks: RoboCasa365~\citep{robocasa365} and LIBERO-Long~\citep{liu2023libero}.
RoboCasa365 Composite-Seen contains 16 tasks in kitchen environments. A 7-DoF Franka Panda arm mounted on a 4-DoF Omron mobile base must coordinate arm manipulation and base navigation. The tasks combine skills such as pick-and-place, articulated-fixture operation, and navigation into extended execution sequences. We evaluate 100 episodes per task across the 10 held-out scenarios.
LIBERO-Long contains 10 tabletop manipulation tasks performed by a fixed-base Franka Panda arm. We evaluate each task on its 50 official initial scenarios. On both benchmarks, we report mean task success rate (SR).

\noindent\textbf{Baselines.}
We report three types of quantitative comparisons. First, we evaluate VLA models after SFT: $\pi_0$~\citep{black2025pi0}, $\pi_{0.5}$~\citep{black2025pi05}, RLDX-1~\citep{kim2026rldx}, and GR00T-N1.5~\citep{bjorck2025gr00t}. We use a benchmark-specific released checkpoint when available; otherwise, we perform SFT following the official recipe.
Second, we compare online VLA RL methods that can be evaluated under a common protocol: Sparse-RL~\citep{zang2026rlinfvla}, SimpleVLA-RL~\citep{li2026simplevla}, and PolicyTrim~\citep{wang2026policytrim}. These methods and {\name} start from the same SFT checkpoints of GR00T-N1.5 and $\pi_{0.5}$. Each method retains its original optimization design and receives the same number of environment interactions. For each benchmark, all online RL methods train one policy jointly across all tasks.
Third, online methods construct dense intermediate rewards using learned progress or value estimators~\citep{shu2025rftf,zhang2026vlac,tan2025robodopamine,liang2026robometer}. We instantiate the Robometer~\citep{liang2026robometer} baseline, a vision-language reward model that predicts per-frame task progress, by integrating its released checkpoint into the RLinf-VLA~\citep{zang2026rlinfvla} training stack and converting its frame-level progress predictions into chunk-level rewards.
Appendix~\ref{app:robometer} details the reward conversion and implementation.

\noindent\textbf{Implementation.}
We implement {\name} in the RLinf-VLA~\citep{zang2026rlinfvla} framework. Before RL training, Claude Opus 4.8~\citep{anthropic2026claudeopus48} generates the subtask decomposition and dependency structure for each task. These decompositions are fixed and reused throughout training, so no LLM is required during rollout collection.
Each retained subtask is then grounded before training to a binary predicate over the simulator state using a fixed rule-based procedure. The verb phrase determines the predicate type, while its arguments identify the task objects or fixtures. We implement these predicates using the benchmark's native state representations and success-check utilities. For example, \emph{place $o$ in $r$} is mapped to a containment predicate that checks whether object $o$ is inside receptacle $r$. Appendix~\ref{app:grounding} provides details on the grounding procedure.
We precompute the reference durations $T_d(v_i)$ for each subtask from the demonstrations used for SFT. Unless otherwise specified, we set $\beta=0.6$ and $\lambda_c=2.0$. The action-chunk length is $L=16$ for GR00T-N1.5 and $L=10$ for $\pi_{0.5}$. All online RL methods are trained for 100 iterations under the same environment-interaction budget. On RoboCasa365, each iteration collects 512 rollouts using two rollout epochs with 256 parallel environments; on LIBERO-Long, each iteration collects 768 rollouts using three rollout epochs with 256 parallel environments. Training uses four nodes with eight NVIDIA A100 GPUs per node and takes approximately 48 hours for RoboCasa365 or 24 hours for LIBERO-Long.

\begin{table*}[!t]
\centering
\caption{
Success rate (\%) by task-horizon bucket. From shortest to longest, the RoboCasa365 buckets contain 3, 5, and 8 tasks, while the LIBERO-Long buckets contain 5, 3, and 2 tasks. \emph{Overall} is the mean SR across all tasks in the benchmark. Within each backbone block, blue cells mark the strongest online RL baseline in each column. $\Delta$ is the difference between {\name} and that baseline, in percentage points.
}
\vspace{-1mm}
\label{tab:horizon-bucket}
\setlength{\tabcolsep}{5pt}
\renewcommand{\arraystretch}{1.15}
\resizebox{\textwidth}{!}{
\begin{tabular}{l | cccc | cccc}
\toprule
\textbf{Benchmark}
 & \multicolumn{4}{c|}{\textbf{RoboCasa365}}
 & \multicolumn{4}{c}{\textbf{LIBERO-Long}} \\
\cmidrule(lr){1-9}
Horizon (steps)
 & 800--1000 & 1000--1400 & 1400--2900 & Overall
 & 250--340 & 340--400 & 400--550 & Overall \\
\midrule
$\pi_0$~\citep{black2025pi0}        &26.0 &23.0 & 8.1&15.5 & 89.6& 90.0& 42.0& 80.2\\
RLDX-1~\citep{kim2026rldx}        &59.0 &55.0 &30.8 &43.6 & 98.0& 94.7& 85.0&94.4 \\
\midrule
GR00T-N1.5~\citep{bjorck2025gr00t}     & 51.7 & 48.0 & 27.9 & 38.6 & 96.0 & 82.0 & 90.0 & 90.6 \\
\quad w/ Sparse-RL~\citep{zang2026rlinfvla}      & \cellcolor{tintblue}60.3 &\cellcolor{tintblue} 52.3 & 25.1 & 40.2 & 92.4 & 90.0 & 90.0 & 91.2 \\
\quad w/ SimpleVLA-RL~\citep{li2026simplevla}   & 55.1 & 50.8 & \cellcolor{tintblue}30.6 & \cellcolor{tintblue}41.5 & \cellcolor{tintblue}94.0 & \cellcolor{tintblue}90.7 & 91.0 & \cellcolor{tintblue}92.4 \\
\quad w/ PolicyTrim~\citep{wang2026policytrim}   & 49.3 & 50.6 & 29.5 & 39.8 & 92.0 & 90.0 & \cellcolor{tintblue}92.0& 91.4 \\
\quad w/ {\name} (\textit{ours})   & 63.0 & 59.0 & 37.6 & 49.1 & 96.0 & 97.3 & 97.0 & 96.6 \\
\rowcolor{tintred}
\quad $\Delta$  & +2.7 & +6.7 & +7.0 & \textbf{+7.6} & +2.0 & +6.6 & +5.0 & \textbf{+4.2} \\
\midrule

$\pi_{0.5}$~\citep{black2025pi05}    & 41.0 & 46.4& 32.9 & 39.3 &94.0 &96.0 &74.0 &90.6 \\
\quad w/ Sparse-RL~\citep{zang2026rlinfvla}      & 43.0 & \cellcolor{tintblue}48.4 & \cellcolor{tintblue}35.8 & 41.1 & \cellcolor{tintblue}97.6 & 94.7 & 77.0 & 92.6 \\
\quad w/ SimpleVLA-RL~\citep{li2026simplevla}    & \cellcolor{tintblue}50.7 & 46.8 & 35.6 & \cellcolor{tintblue}41.9 & 97.2 & \cellcolor{tintblue}96.7 & 82.0 & \cellcolor{tintblue}94.0 \\
\quad w/ PolicyTrim~\citep{wang2026policytrim}   &48.7  &46.0  & 33.5 & 40.3 &  95.6&  90.7&\cellcolor{tintblue}83.0 & 91.6 \\
\quad w/ {\name} (\textit{ours})    & 54.0 & 51.2 & 39.4 & 45.8 & 98.0 & 98.0 & 89.0 & 96.2 \\
\rowcolor{tintred}
\quad $\Delta$  &+3.3  & +2.8 & +3.6 & \textbf{+3.9} & +0.4 & +1.3 & +6.0 & \textbf{+2.2} \\
\bottomrule
\end{tabular}
}
\vspace{-3mm}
\end{table*}

\subsection{Main Results}
Table~\ref{tab:horizon-bucket} reports SR by task-horizon bucket on RoboCasa365 and LIBERO-Long. For GR00T-N1.5, {\name} improves overall SR over the strongest online RL baseline from 41.5\% to 49.1\% on RoboCasa365 and from 92.4\% to 96.6\% on LIBERO-Long, gains of 7.6 and 4.2 percentage points, respectively. For $\pi_{0.5}$, the corresponding improvements are 3.9 percentage points on RoboCasa365 and 2.2 percentage points on LIBERO-Long.

At the bucket level, {\name} exceeds the strongest online RL baseline in every column. The largest positive gains are 7.0 percentage points on RoboCasa365 (1400--2900) and 6.6 percentage points on LIBERO-Long (340--400), both with GR00T-N1.5. These results show that the structured reward improves online RL performance across both evaluated backbones and benchmarks, with substantial gains on several longer-horizon buckets.

\subsection{Reward Source: Structured Events \vs Learned Progress Model}\label{sec:robometer}

\begin{wraptable}[10]{r}{0.45\textwidth}
\vspace{-4pt}
\centering
\caption{
Reward source comparison under a matched PPO training protocol on LIBERO-Long.
}
\vspace{-1mm}
\label{tab:robometer-comparison}
\setlength{\tabcolsep}{4pt}
\renewcommand{\arraystretch}{1.12}
\resizebox{\linewidth}{!}{
\begin{tabular}{lcc}
\toprule
& \multicolumn{2}{c}{\textbf{LIBERO-Long}} \\
\cmidrule(lr){2-3}
Reward Source & GR00T-N1.5 & $\pi_{0.5}$ \\
\midrule
SFT       & 90.6 & 90.6 \\
Robometer~\citep{liang2026robometer} & 94.2 & 93.0 \\
{\name} (\textit{ours})  & $\mathbf{96.6}$ & $\mathbf{96.2}$ \\
\bottomrule
\end{tabular}
}
\vspace{-3pt}
\end{wraptable}

To isolate the effect of reward construction, we compare a Robometer-based reward baseline and {\name} under a matched PPO training setup. Robometer~\citep{liang2026robometer} is a vision-language reward model built on Qwen3-VL-4B~\citep{bai2025qwen3} and trained on 1M trajectories, including LIBERO-Long, to predict scalar per-frame progress. We integrate its released checkpoint into the same RLinf-VLA stack and convert its progress estimates into rewards delivered once per action chunk. For each backbone, both methods use the same SFT initialization, PPO optimizer and hyperparameters, interaction budget, terminal reward, and evaluation protocol. We additionally match Robometer's intermediate-reward scale to that of {\name} on reference rollouts (Appendix~\ref{app:robometer}). Thus, the comparison contrasts learned progress rewards with simulator-verified, dependency-aware completion rewards.

Table~\ref{tab:robometer-comparison} summarizes the results. {\name} outperforms Robometer on both backbones, by 2.4 points with GR00T-N1.5 and 3.2 points with $\pi_{0.5}$. This suggests that simulator-verified structured completion signals provide more effective intermediate supervision than the progress estimates inferred from visual observations. {\name} also does not require a learned reward model or reward model inference during rollout collection.

\begin{figure*}[t]
\begin{center}
\vspace{-2mm}
\includegraphics[width=\linewidth]{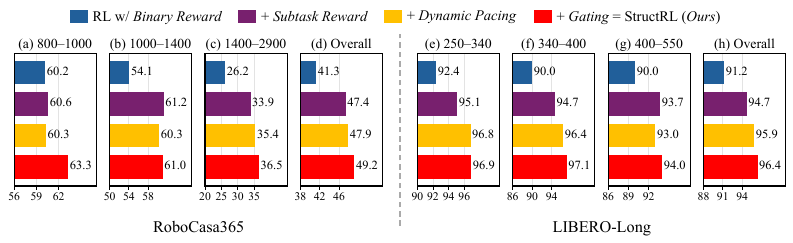}
\end{center}
\vspace{-0.2in}
\caption{
Reward-component ablation with PPO and GR00T-N1.5, adding one component at a time to the terminal binary reward: fixed subtask rewards, dynamic reward pacing, and structure-aware gating, which together form {\name}. Panels report SR (\%) by task-horizon bucket and overall on RoboCasa365 (a--d) and LIBERO-Long (e--h). Values are means over three evaluation runs.
}
\vspace{-0.12in}
\label{fig:ablation}
\end{figure*}

\subsection{Reward-Component Ablation}\label{ablationstudy}
We measure the contribution of each reward component by adding the components one at a time to PPO with GR00T-N1.5. We compare four configurations: (1) a terminal binary reward only; (2) fixed subtask rewards without gating, which pay $\beta/2$, the value of Eq.~\eqref{eq:pace} at $T=T_d$, to each newly detected subtask completion regardless of its prerequisites; (3) dynamic reward pacing, which replaces the fixed reward with the pace-dependent reward in Eq.~\eqref{eq:pace} while remaining ungated; and (4) structure-aware gating applied together with dynamic pacing, yielding the complete {\name} reward. Figure~\ref{fig:ablation} reports SR by task-horizon bucket and overall.

Each added component raises overall SR on both benchmarks. Subtask rewards provide the largest gain, increasing SR from 41.3\% to 47.4\% on RoboCasa365 and from 91.2\% to 94.7\% on LIBERO-Long. This demonstrates the primary benefit of providing intermediate supervision through verifiable subtask completions. Dynamic pacing further improves SR by 0.5 and 1.2 points, respectively, while structure-aware gating adds another 1.3 and 0.5 points, reaching final SRs of 49.2\% and 96.4\%. The complete reward achieves the highest SR in most horizon buckets. Its largest end-to-end gain occurs on the longest RoboCasa365 tasks, where SR increases by 10.3 points, from 26.2\% to 36.5\%.

\subsection{Effect of Subtask Decomposition Density}\label{exp:decomposition}

\begin{wrapfigure}[14]{r}{0.47\textwidth}
\vspace{-4pt}
\centering
\includegraphics[width=\linewidth]{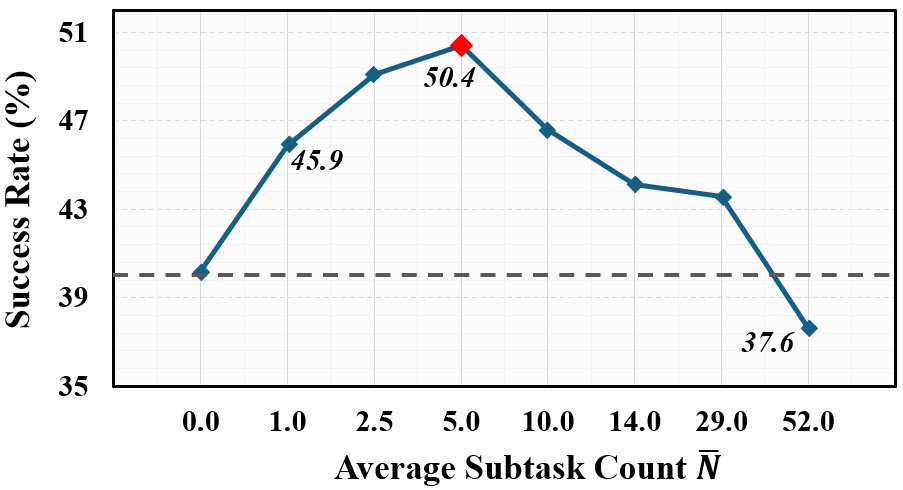}
\vspace{-4pt}
\caption{
Effect of subtask decomposition density on RoboCasa365 using GR00T-N1.5.
}
\label{fig:ablation-density}
\vspace{-4pt}
\end{wrapfigure}

Finer decompositions provide more opportunities for intermediate supervision, but adding more completion signals may not always improve learning. We study this tradeoff using GR00T-N1.5 on the 16 RoboCasa365 tasks. Let $\bar{N}$ denote the average number of subtask completion signals per task. We evaluate eight settings, from $\bar{N}=0$, which uses only the terminal binary reward, to increasingly fine-grained decompositions generated by the LLM.
For this controlled sweep, we relax the default filtering criteria and permit fine motion checkpoints that would normally be excluded as weak indicators of task progress. Every retained checkpoint still has an executable binary predicate so that it can be evaluated during RL.
We train a separate policy for each setting and report its overall SR in Figure~\ref{fig:ablation-density}. Appendix~\ref{app:decomposition-density} illustrates the decomposition ladder.

Intermediate supervision helps even at low density: increasing $\bar{N}$ from 0 to 1 raises SR from 40.2\% to 45.9\%. The benefit is not monotonic, however. Performance reaches 50.4\% at $\bar{N}=5$ and then declines as the decomposition becomes finer. At $\bar{N}=52$, SR falls to 37.6\%, below the terminal-only configuration. Thus, within this controlled sweep, a moderate decomposition density is more effective than either terminal-only reward or an excessively fine decomposition. 
Controlled analyses in Appendix~\ref{app:deccline_analysis} indicate that the decline reflects both earlier, less task-aligned completion signals and growth in the total intermediate reward as more signals are added.

\begin{table}[!t]
\centering

\begin{minipage}[t]{0.45\textwidth}
\vspace{0pt}
\centering

\captionof{table}{
Effect of the policy optimizer on RoboCasa365. We train {\name} with PPO or GRPO on GR00T-N1.5 and $\pi_{0.5}$ and report SR (\%) by task-horizon bucket. SimpleVLA-RL is included for reference.
}
\label{tab:ablation_grpo}

\setlength{\tabcolsep}{5pt}
\renewcommand{\arraystretch}{1.15}

\resizebox{\linewidth}{!}{%
\begin{tabular}{l | ccc | c}
\toprule
Method
 & 800--1000 & 1000--1400 & 1400--2900 & Overall \\
\midrule
\multicolumn{5}{c}{\textit{GR00T-N1.5}} \\
\quad w/ SimpleVLA-RL & 55.1 & 50.8 & 30.6 & 41.5 \\
\quad w/ {\name}        & 63.0 & 59.0 & 37.6 & \textbf{49.1} \\
\rowcolor{tintgray}
\quad w/ {\name}-GRPO   & 61.3 & 55.4 & 32.2 & 44.9 \\
\midrule
\multicolumn{5}{c}{\textit{$\pi_{0.5}$}} \\
\quad w/ SimpleVLA-RL & 50.7 & 46.8 & 35.6 & 41.9 \\
\quad w/ {\name}        & 54.0 & 51.2 & 39.4 & \textbf{45.8} \\
\rowcolor{tintgray}
\quad w/ {\name}-GRPO   & 51.0 & 48.4 & 37.0 & 43.2 \\
\bottomrule
\end{tabular}%
}

\end{minipage}
\hfill
\begin{minipage}[t]{0.53\textwidth}
\centering

\captionof{table}{
Zero-shot evaluation on additional RoboCasa365 suites using GR00T-N1.5. Policies from Table~\ref{tab:horizon-bucket} are evaluated without further training on 16 held-out long-horizon tasks from Composite-Unseen and 65 atomic tasks from Atomic-Seen.
}
\label{tab:cross-test-avg}

\setlength{\tabcolsep}{6pt}
\renewcommand{\arraystretch}{1.2}

\resizebox{\linewidth}{!}{%
\begin{tabular}{l | cc | c}
\toprule
\multirow{2}{*}{Method}
 & Composite & Atomic & \multirow{2}{*}{\textbf{Average}} \\
 & Unseen & Seen & \\
\midrule
GR00T-N1.5
& 3.5
& \cellcolor{tintblue}17.0
& 14.4 \\

\quad w/ SimpleVLA-RL
& \cellcolor{tintblue}4.3
& 16.9
& \cellcolor{tintblue}14.5 \\

\quad w/ {\name} (\textit{ours})
& 4.8
& 20.5
& 17.6 \\

\rowcolor{tintred}
\quad $\Delta$
& +0.5
& +3.5
& \textbf{+3.1} \\
\bottomrule
\end{tabular}%
}

\end{minipage}

\vspace{-4pt}
\end{table}

\subsection{Compatibility with GRPO}\label{sec:extend_grpo}

We evaluate whether the structured reward remains effective when PPO is replaced by GRPO. On RoboCasa365, we train GR00T-N1.5 and $\pi_{0.5}$ using the rollout-level formulation from Section~\ref{sec:method_grpo}. Table~\ref{tab:ablation_grpo} reports the results. {\name}-GRPO exceeds SimpleVLA-RL in every horizon bucket, improving overall SR by 3.4 percentage points with GR00T-N1.5 and 1.3 percentage points with $\pi_{0.5}$. This result shows that the structured reward is compatible with the evaluated GRPO formulation. The PPO variant remains stronger overall, by 4.2 and 2.6 percentage points on the two backbones, respectively, supporting its use as the default optimizer. One possible explanation is the difference in credit granularity: PPO associates each reward with the action chunk that triggers it, whereas GRPO aggregates all chunk-level rewards into one rollout return before computing group-relative advantages.

\subsection{Zero-Shot Evaluation on Additional RoboCasa365 Tasks}
We evaluate the GR00T-N1.5 policies from Table~\ref{tab:horizon-bucket} without further training on two RoboCasa365 suites excluded from RL training: 16 held-out long-horizon tasks from Composite-Unseen and 65 atomic tasks from Atomic-Seen. Table~\ref{tab:cross-test-avg} reports both evaluations.

On Composite-Unseen, SR remains low for all evaluated policies. {\name} reaches 4.8\%, compared with 4.3\% for the strongest non-{\name} comparison. These results indicate limited zero-shot transfer to unseen task compositions under the current setup.

Atomic-Seen provides a complementary test of forgetting: these tasks were encountered during pretraining but excluded from composite-task RL. As shown in Table~\ref{tab:cross-test-avg}, {\name} reaches 20.5\% SR, compared with 17.0\% for the SFT policy and 16.9\% for SimpleVLA-RL.
Thus, the composite-task RL update does not reduce Atomic-Seen performance under {\name}; instead, the evaluated checkpoint improves over SFT. One possible explanation is that intermediate completion rewards reinforce skills shared across atomic and composite tasks, although this table does not isolate that mechanism.

\subsection{Applicability to Shorter-Horizon Tasks}
Structured intermediate rewards are motivated by the difficulty of assigning credit over long horizons. We therefore ask whether the same framework remains useful when tasks are shorter and baseline performance is already high. We evaluate {\name} on the LIBERO Spatial, Object, and Goal suites using GR00T-N1.5, training a separate policy on each suite with the same decomposition and optimization procedures used for LIBERO-Long. We make no short-task-specific modification; for example, a pick-and-place task can provide separate completion signals for grasping the object and placing it at the target.

\begin{wraptable}[11]{r}{0.47\textwidth}
\vspace{-4pt}
\centering
\caption{
Success rate (\%) on the LIBERO Spatial, Object, and Goal suites using GR00T-N1.5. A separate policy is trained on each suite using the same optimization recipe.
}
\label{tab:libero-suites}

\setlength{\tabcolsep}{6pt}
\renewcommand{\arraystretch}{1.2}

\resizebox{\linewidth}{!}{%
\begin{tabular}{l | ccc | c}
\toprule
Method & Spatial & Object & Goal & \textbf{Average} \\
\midrule
GR00T-N1.5
& 89.3
& 97.9
& 94.6
& 93.7 \\

\quad w/ SimpleVLA-RL
& \cellcolor{tintblue}91.2
& \cellcolor{tintblue}98.5
& \cellcolor{tintblue}94.8
& \cellcolor{tintblue}94.7 \\

\quad w/ {\name} (\textit{ours})
& 92.1
& 99.4
& 95.6
& 95.7 \\

\rowcolor{tintred}
\quad $\Delta$
& +0.9
& +0.9
& +0.8
& \textbf{+1.0} \\
\bottomrule
\end{tabular}%
}
\vspace{-3pt}
\end{wraptable}

As shown in Table~\ref{tab:libero-suites}, {\name} improves over SimpleVLA-RL from 91.2\% to 92.1\% on Spatial, from 98.5\% to 99.4\% on Object, and from 94.8\% to 95.6\% on Goal. These gains are modest, consistent with the high baseline success rates, but positive on all three suites. The results show that the same structured-reward pipeline remains effective when task structure is shallower, while the larger benefits remain concentrated in the long-horizon setting for which the method is designed.

\noindent\textbf{Additional details and experimental results.}
The appendix provides the complete training algorithm (Appendix~\ref{app:training}); decomposition details and decomposer sensitivity (Appendices~\ref{app:decomp_prompt} and~\ref{app:LLM_decomp}); density, reward-mechanism, and pacing analyses (Appendices~\ref{app:density-ladder}--\ref{app:demofree-pacing}); repeated evaluations, training dynamics, and reward sensitivity (Appendices~\ref{app:random}--\ref{app:hyper}); case studies (Appendix~\ref{app:case_study}); and Robometer implementation details (Appendix~\ref{app:robometer}).

\section{Related Work}\label{sec:related_work}

\noindent\textbf{Long-horizon VLA tasks.}
VLA models have achieved strong performance on conventional manipulation tasks, including individual pick-and-place skills~\citep{black2025pi05,bjorck2025gr00t,physicalintelligence2025pi06}. Recent benchmarks extend evaluation to tasks that compose multiple skills, using either physical robots~\citep{wu2025robomind,fu2024mobilealoha} or reproducible simulators~\citep{liu2023libero,han2025robocerebra,robocasa365}. For example, storing leftovers in RoboCasa365 requires the policy to sort food into containers, carry them across the kitchen, and place them in a refrigerator. Such tasks jointly test manipulation, instruction following, navigation, and extended execution. Because these policies are commonly trained through SFT on demonstrations, their performance remains sensitive to demonstration coverage and compounding execution errors. {\name} uses online interaction and intermediate rewards to improve these policies after SFT.

\noindent\textbf{Reinforcement learning for VLAs.}
RL is increasingly used to post-train VLA policies. Offline approaches optimize from pre-collected rollouts~\citep{zhang2024grape,zhang2025reinbot,chen2025conrft,huang2025corft}, while recent online methods collect new interactions and optimize with PPO~\citep{zang2026rlinfvla,lu2025vlarl,liu2025rlvla,chen2025pirl} or GRPO~\citep{li2026simplevla,wang2026policytrim,chen2025tgrpo,tan2025riptvla}. Many online methods provide reward only after the complete rollout succeeds, making the signal increasingly sparse as the execution horizon grows.
A complementary line of work provides denser supervision using learned progress or value estimators~\citep{shu2025rftf,zhang2026vlac,tan2025robodopamine,liang2026robometer}.
However, these estimators may require a large model during rollout collection or task-specific reward-model training.
{\name} derives intermediate rewards directly from simulator-verifiable subtask completions and structures them with prerequisite dependencies and demonstration-derived pacing, eliminating the need for a learned reward model during rollout collection.

\section{Conclusion}
We presented {\name}, an online RL framework that replaces terminal-only supervision with structured intermediate rewards for long-horizon VLA tasks. {\name} decomposes each instruction into verifiable subtasks, uses their dependencies to determine when a completion constitutes valid progress, and scales the resulting reward according to completion pace. Across RoboCasa365 and LIBERO-Long, {\name} improves GR00T-N1.5 and $\pi_{0.5}$ over the evaluated online RL baselines, with the largest gains on longer-horizon tasks for GR00T-N1.5.
These results show that verifiable subtask rewards, organized by task structure and calibrated to demonstration pace, can improve long-horizon VLA post-training.

\subsection*{AI use statement}
We used generative AI tools to assist with experimental infrastructure, log analysis, and drafting and revising manuscript text. The authors independently verified all reported results, numerical values, and claims and take full responsibility for the paper.

\subsection*{Ethics statement}
Online VLA RL requires substantial computation. A typical run uses 32 NVIDIA A100 GPUs for approximately 48 hours on RoboCasa365 or 24 hours on LIBERO-Long, corresponding to about 1,536 and 768 GPU-hours, respectively.

Shaped rewards can induce unintended behavior. In some rollouts, policies collected intermediate rewards and then remained idle until timeout. Dynamic reward pacing discourages delaying future subtask completions because the available reward decreases with completion time, but it does not eliminate idling after intermediate rewards have already been collected. Other reward-induced failure modes may therefore remain. Because {\name} favors faster subtask completion, deployment on physical robots would require separate safety validation.

\subsection*{Reproducibility Statement}
We provide the training and evaluation code, generated subtask decompositions, and reward implementation at \url{\codeurl}. The main text and appendix specify the reward formulation, the reward hyperparameters, action-chunk lengths, reference completion durations, and interaction budget. We release the generated subtask graphs as static configuration files, allowing the experiments to be reproduced without querying an LLM or depending on a particular model version.

Evaluation follows each benchmark’s official task horizons and initial-state protocol. RoboCasa365 uses 16 composite tasks with 100 episodes per task, while LIBERO-Long uses 10 tasks with 50 episodes per task. We compute macro SR as the mean of per-task success rates and state the number of repeated runs wherever a standard deviation is reported.

\bibliographystyle{iclr2027_conference}
\bibliography{iclr2027_conference}

\clearpage
\appendix

\addtocontents{toc}{\protect\setcounter{tocdepth}{2}}
\definecolor{linkcolor}{HTML}{000000}
\newpage
\section*{\centering\LARGE Appendix}
\tableofcontents
\newpage
\appendix
\definecolor{linkcolor}{HTML}{ED1C24}

\section{Training Procedure}
\label{app:training}
Algorithm~\ref{alg:training} states the full training loop of {\name} in one
place: rollout collection, the structure-aware reward gating and dynamic reward pacing described in Section~\ref{sec:method}, and the PPO update.

\begin{algorithm}[!ht]
\small
\caption{Overall Training Process of {\name}}
\label{alg:training}
\begin{algorithmic}[1]
\Require SFT-initialized VLA policy $\pi_\theta$; decomposition
$\{(v_i,X(v_i))\}_{i=1}^{N}$ from Eq.~\eqref{eq:decomp}; demonstration
durations $\{T_d(v_i)\}_{i=1}^{N}$; hyperparameters $\beta$, $\lambda_c$
\For{each training iteration}
\State \textbf{\textcolor{red}{// Step 1: Rollout collection}}
\State Roll out $\pi_\theta$ to collect $\tau=(o_1,a_1,\dots,o_M,a_M)$
\State \textbf{\textcolor{red}{// Step 2: Structured reward computation}}
\State $\mathcal{C}\gets\varnothing$;\; $k_{\text{last}}\gets 0$;\; $r_k\gets 0$ for all $k$
\For{$k=1,\dots,M$}
    \State \textcolor{gray}{\# Reward Gating}
    \ForAll{$v_i\notin\mathcal{C}$ with $X(v_i)\subseteq\mathcal{C}$}
        \If{$v_i$ is completed during $a_k$}
            \State $T(v_i)\gets k-k_{\text{last}}$
            \State \textcolor{gray}{\# Reward Pacing}
            \State $r_k \gets r_k + \beta/\big(1+T(v_i)/T_d(v_i)\big)$
            \State $\mathcal{C}\gets\mathcal{C}\cup\{v_i\}$;\;\; $k_{\text{last}}\gets k$
        \EndIf
    \EndFor
\EndFor
\State $r_M \gets r_M + \lambda_c$ \textbf{if} the task is completed at $a_M$
\State \textbf{\textcolor{red}{// Step 3: Policy optimization}}
\State Update $\pi_\theta$ via PPO with $\tau$ and $\{r_k\}_{k=1}^{M}$ \Comment{Eq.~\eqref{eq:ppo}}
\EndFor
\State \Return optimized VLA policy $\pi_\theta$
\end{algorithmic}
\end{algorithm}

\section{Subtask Decomposition}\label{app:decomp_prompt}

\subsection{LLM Decomposition Prompt}
We provide the prompt used to decompose each long-horizon task (Section~\ref{sec:decomp}).
The LLM receives only the requirements below and the benchmark's natural-language task command. It is given no simulator predicate vocabulary, demonstration data, or example decomposition, so the proposal depends only on the task command. Line breaks inside the prompt are for typesetting only.

{\footnotesize
\begin{verbatim}
You are a robot task planner. Decompose a kitchen manipulation
task (performed by a single-arm mobile robot) into stages of
subtasks.

Requirements:
1. Output STAGES in strict execution order: stage k+1 can only
   start after every subtask in stage k is done.
2. Subtasks WITHIN one stage may be completed in any order
   (parallel set).
3. Each subtask must be ONE atomic, physically checkable
   manipulation event, phrased as a short verb phrase in this
   controlled form:
   - "grasp <object>"
   - "place <object> in/on <receptacle or location>"
   - "open <fixture>" / "close <fixture>"
   - "turn on <fixture>" / "turn off <fixture>"
   - "press <button>"
   - "<activity> for a while" for continuous activities
     (stirring, washing, scrubbing), optionally split into
     progressive milestones.
4. Include intermediate manipulation events (like grasping an
   object before placing it), not just final outcomes.
5. Do NOT include a final "task finished" subtask, and do NOT
   include robot retreat/release-and-back-away steps. Each retained
   subtask should correspond to a meaningful state transition that
   reflects genuine progress toward the task goal.
6. Output ONLY a JSON object:
   {"stages": [["subtask", ...], ...]} - no prose.
\end{verbatim}
}

\noindent The user turn is the single line \texttt{Task command: "\{cmd\}"} followed by \texttt{Decompose this task now.}, where \texttt{\{cmd\}} is the benchmark command. Requirement~3 is the \emph{verifiability} constraint of Section~\ref{sec:decomp} expressed as a controlled output form, and requirement~5 excludes the two classes of non-progress event that constitute the \emph{progress} constraint. Decoding is greedy (\texttt{do\_sample=False}), so a proposal is reproducible given the model. Table~\ref{tab:decomp-released} lists the grounded decompositions used in RoboCasa365 experiments.

\subsection{Grounding Subtasks to Simulator Predicates}\label{app:grounding}

Before RL training, each retained natural-language subtask is deterministically grounded to a binary predicate over simulator state. Grounding is performed once per task, and the resulting predicates are fixed throughout training.

\noindent\textbf{RoboCasa365 grounding.}
We implement predicates using the benchmark's simulator state and native task-success utilities, with the same thresholds where applicable. Each supported verb form maps to a predicate template, while its arguments bind the template to the relevant task objects or fixtures. Table~\ref{tab:grounding-templates} summarizes the mappings.

\begin{table}[!ht]
\centering
\vspace{-3mm}
\caption{Predicate templates for grounding RoboCasa365 subtasks. $o$, $r$, and $f$ denote a bound object, receptacle, and fixture.}
\vspace{-2mm}
\label{tab:grounding-templates}
\small
\setlength{\tabcolsep}{4pt}
\renewcommand{\arraystretch}{1.1}
\begin{tabular}{@{}l p{0.52\textwidth} l@{}}
\toprule
Subtask phrase & Predicate is true when & Example \\
\midrule
place $o$ in/on $r$
& $o$ is contained in $r$ or rests on it
& \texttt{chicken\_in\_bowl} \\

grasp $o$
& $o$ is held by the gripper; once detected, the predicate is latched
& \texttt{grasp\_straw} \\

open / close $f$
& the relevant door or drawer crosses the benchmark threshold
& \texttt{dishwasher\_closed} \\

turn on / off $f$, press $f$
& $f$ reaches the corresponding discrete state
& \texttt{water\_on} \\

$\langle$activity$\rangle$ for a while
& a simulator-maintained activity timer reaches a specified threshold
& \texttt{wash\_t10} \\
\bottomrule
\end{tabular}
\end{table}

Object and fixture mentions are bound to the simulator handles associated with the task command. When multiple instances of the same category occur, repeated mentions are assigned to the corresponding task instances.

\noindent\textbf{Filtering.}
We discard grounded subtasks if (i) no supported predicate template exists, (ii) the predicate is already satisfied at reset or does not reliably indicate task progress, (iii) it cannot be reliably timed from the SFT demonstrations, or (iv) it duplicates another predicate. For timing, we replay the 100 demonstrations for each task and remove predicates that trigger too infrequently to estimate $T_d$ or whose relevant state is not reproduced under replay. The LLM-provided dependency order is preserved after filtering, with empty groups removed.

\noindent\textbf{LIBERO.}
On LIBERO, we directly use the goal conjuncts from each task's BDDL specification, evaluated by the benchmark's native goal checker. For example, \emph{put the black bowl in the bottom drawer of the cabinet and close it} yields an \texttt{In} predicate for the bowl and drawer, followed by a \texttt{Close} predicate for the drawer. Goal conjuncts already satisfied at reset are removed. For the Spatial, Object, and Goal suites, we additionally insert a grasp predicate before each placement predicate; articulated, pushing, and knob tasks retain a single goal group.

\begin{table}[!ht]
\centering
\caption{
Subtask decompositions for the 16 RoboCasa365 composite-seen tasks at
$\bar{N}=2.375$ and $5.0$.
Braces group unordered subtasks within a stage, while arrows indicate strict
stage ordering. $N$ denotes the number of progress signals.
The $\bar{N}=5.0$ setting additionally includes grasp milestones,
intermediate fixture states, and timed activity checkpoints.
Retreat signals are appended automatically and excluded from $N$.
}
\label{tab:decomp-released}

\scriptsize
\setlength{\tabcolsep}{2pt}
\renewcommand{\arraystretch}{0.90}

\newcolumntype{L}[1]{>{\raggedright\arraybackslash}p{#1}}

\begin{tabular}{
@{}
L{0.135\textwidth}
r
L{0.315\textwidth}
r
L{0.465\textwidth}
@{}
}
\toprule
\textbf{Task}
& \multicolumn{2}{c}{\textbf{$\bar{N}=2.375$}}
& \multicolumn{2}{c}{\textbf{$\bar{N}=5.0$}}
\\
\cmidrule(lr){2-3}
\cmidrule(l){4-5}
& $N$ & \textbf{Progress signals}
& $N$ & \textbf{Progress signals}
\\
\midrule

\textit{DeliverStraw}
& 1
& \{\texttt{straw\_in\_glass\_cup}\}
& 2
& \{\texttt{grasp\_straw}\}
  \stagearrow
  \{\texttt{straw\_in\_glass\_cup}\}
\\

\textit{GetToastedBread}
& 2
& \{\texttt{toaster\_on}\}
  \stagearrow
  \{\texttt{toast\_on\_plate}\}
& 3
& \{\texttt{toaster\_on}\}
  \stagearrow
  \{\texttt{grasp\_toast}\}
  \stagearrow
  \{\texttt{toast\_on\_plate}\}
\\

\textit{KettleBoiling}
& 2
& \{\texttt{kettle\_on\_stove}\}
  \stagearrow
  \{\texttt{kettle\_on\_active\_burner}\}
& 3
& \{\texttt{grasp\_kettle}\}
  \stagearrow
  \{\texttt{kettle\_on\_stove}\}
  \stagearrow
  \{\texttt{kettle\_on\_active\_burner}\}
\\

\textit{LoadDishwasher}
& 3
& \{\texttt{dish0\_on\_rack}, \texttt{dish1\_on\_rack}\}
  \stagearrow
  \{\texttt{dishwasher\_closed}\}
& 6
& \{\texttt{grasp\_dish0}, \texttt{dish0\_on\_rack},
  \texttt{grasp\_dish1}, \texttt{dish1\_on\_rack}\}
  \stagearrow
  \{\texttt{door\_half\_closed}\}
  \stagearrow
  \{\texttt{dishwasher\_closed}\}
\\

\textit{PackIdenticalLunches}
& 2
& \{\texttt{tupper0\_complete}, \texttt{tupper1\_complete}\}
& 8
& \{\texttt{grasp\_vegetable0}, \texttt{vegetable0\_packed},
  \texttt{grasp\_vegetable1}, \texttt{vegetable1\_packed},
  \texttt{grasp\_meat0}, \texttt{meat0\_packed},
  \texttt{grasp\_meat1}, \texttt{meat1\_packed}\}
\\

\textit{PreSoakPan}
& 3
& \{\texttt{water\_on}, \texttt{pan\_in\_sink},
  \texttt{sponge\_in\_sink}\}
& 5
& \{\texttt{grasp\_pan}, \texttt{pan\_in\_sink},
  \texttt{grasp\_sponge}, \texttt{sponge\_in\_sink},
  \texttt{water\_on}\}
\\

\textit{PrepareCoffee}
& 2
& \{\texttt{mug\_at\_machine}\}
  \stagearrow
  \{\texttt{machine\_turned\_on}\}
& 3
& \{\texttt{grasp\_mug}\}
  \stagearrow
  \{\texttt{mug\_at\_machine}\}
  \stagearrow
  \{\texttt{machine\_turned\_on}\}
\\

\textit{RinseSinkBasin}
& 3
& \{\texttt{washed\_left}, \texttt{washed\_center},
  \texttt{washed\_right}\}
& 4
& \{\texttt{water\_on}\}
  \stagearrow
  \{\texttt{washed\_left}, \texttt{washed\_center},
  \texttt{washed\_right}\}
\\

\textit{ScrubCuttingBoard}
& 2
& \{\texttt{contact\_5steps}, \texttt{sweep\_range\_0p1m}\}
& 6
& \{\texttt{grasp\_sponge}\}
  \stagearrow
  \{\texttt{contact\_1step}\}
  \stagearrow
  \{\texttt{contact\_3steps}, \texttt{sweep\_range\_0p05m}\}
  \stagearrow
  \{\texttt{contact\_5steps}, \texttt{sweep\_range\_0p1m}\}
\\

\textit{SearingMeat}
& 2
& \{\texttt{meat\_in\_pan}\}
  \stagearrow
  \{\texttt{pan\_on\_active\_knob}\}
& 5
& \{\texttt{grasp\_pan}, \texttt{pan\_on\_stove},
  \texttt{pan\_on\_active\_knob}\}
  \stagearrow
  \{\texttt{grasp\_meat}, \texttt{meat\_in\_pan}\}
\\

\textit{SetUpCuttingStation}
& 2
& \{\texttt{meat\_on\_board}, \texttt{knife\_on\_board}\}
& 4
& \{\texttt{grasp\_meat}, \texttt{meat\_on\_board},
  \texttt{grasp\_knife}, \texttt{knife\_on\_board}\}
\\

\textit{StackBowlsCabinet}
& 2
& \{\texttt{bowls\_stacked}\}
  \stagearrow
  \{\texttt{any\_bowl\_in\_cabinet}\}
& 4
& \{\texttt{grasp\_any\_bowl}\}
  \stagearrow
  \{\texttt{bowls\_stacked}\}
  \stagearrow
  \{\texttt{any\_bowl\_in\_cabinet},
  \texttt{both\_bowls\_in\_cabinet}\}
\\

\textit{SteamInMicrowave}
& 3
& \{\texttt{veg\_in\_bowl}\}
  \stagearrow
  \{\texttt{bowl\_in\_micro}\}
  \stagearrow
  \{\texttt{door\_closed}\}
& 6
& \{\texttt{grasp\_vegetable}, \texttt{veg\_in\_bowl}\}
  \stagearrow
  \{\texttt{grasp\_bowl}, \texttt{bowl\_in\_micro}\}
  \stagearrow
  \{\texttt{door\_half\_closed}, \texttt{door\_closed}\}
\\

\textit{StirVegetables}
& 4
& \{\texttt{veg1\_in\_pot}, \texttt{veg2\_in\_pot}\}
  \stagearrow
  \{\texttt{spatula\_grasped}\}
  \stagearrow
  \{\texttt{task\_complete}\}
& 8
& \{\texttt{grasp\_veg1}, \texttt{veg1\_in\_pot},
  \texttt{grasp\_veg2}, \texttt{veg2\_in\_pot}\}
  \stagearrow
  \{\texttt{spatula\_grasped}\}
  \stagearrow
  \{\texttt{stir\_t1}, \texttt{stir\_t3}\}
  \stagearrow
  \{\texttt{task\_complete}\}
\\

\textit{StoreLeftoversInBowl}
& 3
& \{\texttt{chicken\_in\_bowl}, \texttt{vegetable\_in\_bowl}\}
  \stagearrow
  \{\texttt{bowl\_in\_fridge}\}
& 6
& \{\texttt{grasp\_chicken}, \texttt{chicken\_in\_bowl},
  \texttt{grasp\_vegetable}, \texttt{vegetable\_in\_bowl}\}
  \stagearrow
  \{\texttt{grasp\_bowl}, \texttt{bowl\_in\_fridge}\}
\\

\textit{WashLettuce}
& 2
& \{\texttt{water\_on}\}
  \stagearrow
  \{\texttt{task\_complete}\}
& 7
& \{\texttt{water\_on}\}
  \stagearrow
  \{\texttt{lettuce\_under\_water}\}
  \stagearrow
  \{\texttt{wash\_t5}, \texttt{wash\_t10},
  \texttt{wash\_t15}, \texttt{wash\_t20}\}
  \stagearrow
  \{\texttt{task\_complete}\}
\\

\midrule
\textbf{Total}
& 38 & 
& 80 & \\
\bottomrule
\end{tabular}
\end{table}

\subsection{Sensitivity to the Decomposer}\label{app:LLM_decomp}
Because the decomposition is authored by an LLM, we test sensitivity to the model choice. Table~\ref{tab:ablation-decomp-llm} compares the default decomposer (Claude Opus 4.8) against a weaker open-weight model, Qwen3.5-9B~\cite{qwen35blog}, under identical training hyperparameters.
Replacing Claude Opus 4.8 with Qwen3.5-9B reduces overall SR from 49.1\% to 45.7\%. This result shows that the pipeline can be instantiated with the evaluated open-weight decomposer, while the remaining gap indicates that decomposition quality affects downstream RL performance.

\begin{table}[!ht]
\centering
\caption{Ablation on the LLM that authors the subtask decomposition. Given
only the task language command and our decomposition requirements, each LLM
proposes the stage/subtask split; {\name} is then trained with identical
hyperparameters. SR (\%) grouped by horizon bucket. In Table~\ref{tab:horizon-bucket}, we use Claude Opus 4.8 by default for decomposition.}
\label{tab:ablation-decomp-llm}
\setlength{\tabcolsep}{5pt}
\renewcommand{\arraystretch}{1.2}
\resizebox{0.6\columnwidth}{!}{%
\begin{tabular}{l | ccc | c}
\toprule
Horizon (steps)
 & 800--1000 & 1000--1400 & 1400--2900 & Overall\\
\midrule
Opus 4.8 & 63.0 & 59.0 & 37.6 & 49.1 \\
Qwen3.5-9B & 59.3 &60.6  & 31.2 & 45.7 \\
\bottomrule
\end{tabular}%
}
\end{table}

\section{Further Analysis of Subtask Decomposition Granularity}\label{app:decomposition-density}

Section~\ref{exp:decomposition} shows that performance peaks at moderate decomposition density and declines when the decomposition becomes excessively fine. This appendix details the density construction, reports results by horizon bucket, and analyzes two contributors to the decline. All experiments use GR00T-N1.5 on RoboCasa365 with fixed training hyperparameters unless stated otherwise.

\subsection{Constructing the Density Ladder}
\label{app:density-ladder}

\begin{figure}[!t]
\begin{center}

\includegraphics[width=0.95\linewidth]{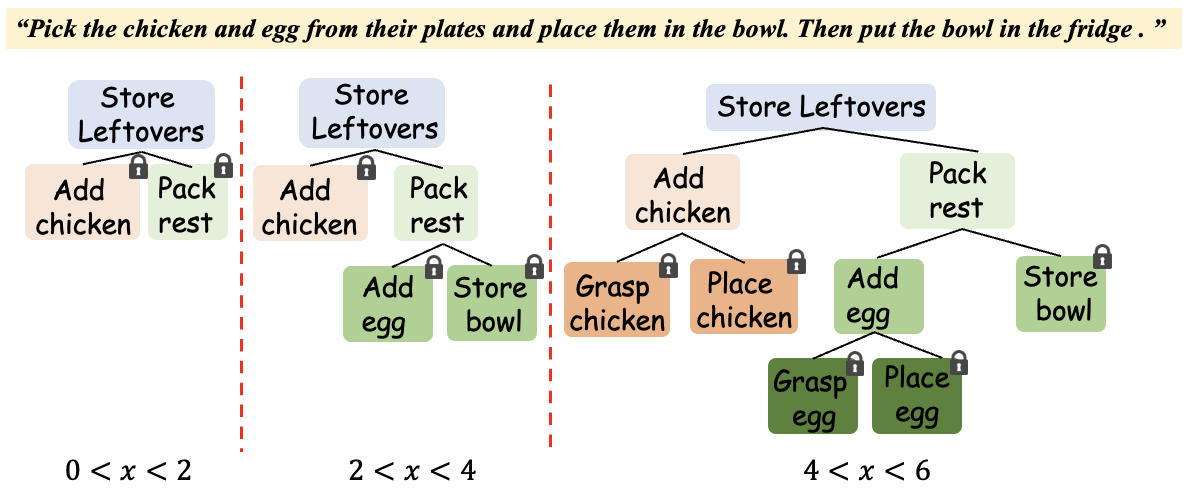}
\end{center}
\vspace{-0.1in}
\caption{An example of hierarchical subtask decomposition under increasing reward density.
Each level refines the previous tree by splitting parent subtasks into finer
children while keeping all existing milestones unchanged; \underline{\emph{lock icons}} mark the
subtasks whose completion will be detected and rewarded during RL training at the
corresponding density level.}

\label{fig:split_tree}
\end{figure}

We first illustrate how a single task is progressively decomposed into subtask sets with different density levels. As shown in Figure~\ref{fig:split_tree}, successive density levels are constructed by refining each existing milestone into finer-grained child subtasks while keeping all previously introduced milestones intact, thereby progressively increasing the number of subtasks. During this process, we remove the verifiability and progress constraints used in our default subtask decomposition procedure, allowing the LLM to decompose each task as finely as possible. Finally, Table~\ref{tab:reward_density_example} provides one representative task and lists the specific subtasks corresponding to each density level for reference.

\begin{table}[!t]
\centering
\caption{
Reward decomposition for \textit{StoreLeftoversInBowl} at different densities.
Here, $\bar{N}$ is the average number of gated subtask signals. The first four levels correspond
to the trees in Fig.~\ref{fig:split_tree}; denser levels progressively refine the
same six milestones with finer verifiable motion checkpoints.
}
\small
\begin{tabular}{@{}c p{0.84\textwidth}@{}}
\toprule
\textbf{Density $\bar{N}$} & \textbf{Rewarded subtasks for \textit{StoreLeftoversInBowl}} \\
\midrule

$0.0$ &
Terminal reward only. \\[2pt]

$1.0$ &
Chicken in bowl. \\[2pt]

$2.5$ &
Chicken in bowl; vegetable in bowl; bowl in fridge. \\[2pt]

$5.0$ &
Grasp chicken; chicken in bowl; grasp vegetable; vegetable in bowl;
grasp bowl; bowl in fridge. \\[2pt]

$10.0$ &
Refine the six milestones with intermediate motion checkpoints:
reach/approach before each grasp, and lift/carry/near-fridge before placement. \\[2pt]

$14.0$ &
Further refine them with gripper closing, above-the-bowl checkpoints,
bowl contact, and additional stages of the fridge approach. \\[2pt]

$29.0$ &
Represent each milestone as a short chain of 5–6 checks, including staged
distance reduction, contact, gripper closure, stable grasp, lift, and staged
approach to the receptacle. \\[2pt]

$52.0$ &
Use 10–11 checkpoints per milestone, covering progressively shrinking
gripper–object distances, contact, finger closure, stable grasp, lift,
receptacle approach, above, lowered, inside, and released states. \\

\bottomrule
\end{tabular}

\label{tab:reward_density_example}
\end{table}

\begin{figure}[!ht]
\centering

\begin{minipage}[t]{0.43\textwidth}
\vspace{0pt}
\centering
\captionof{figure}{SR trends across decomposition densities for different horizon buckets.}
\includegraphics[width=\linewidth]{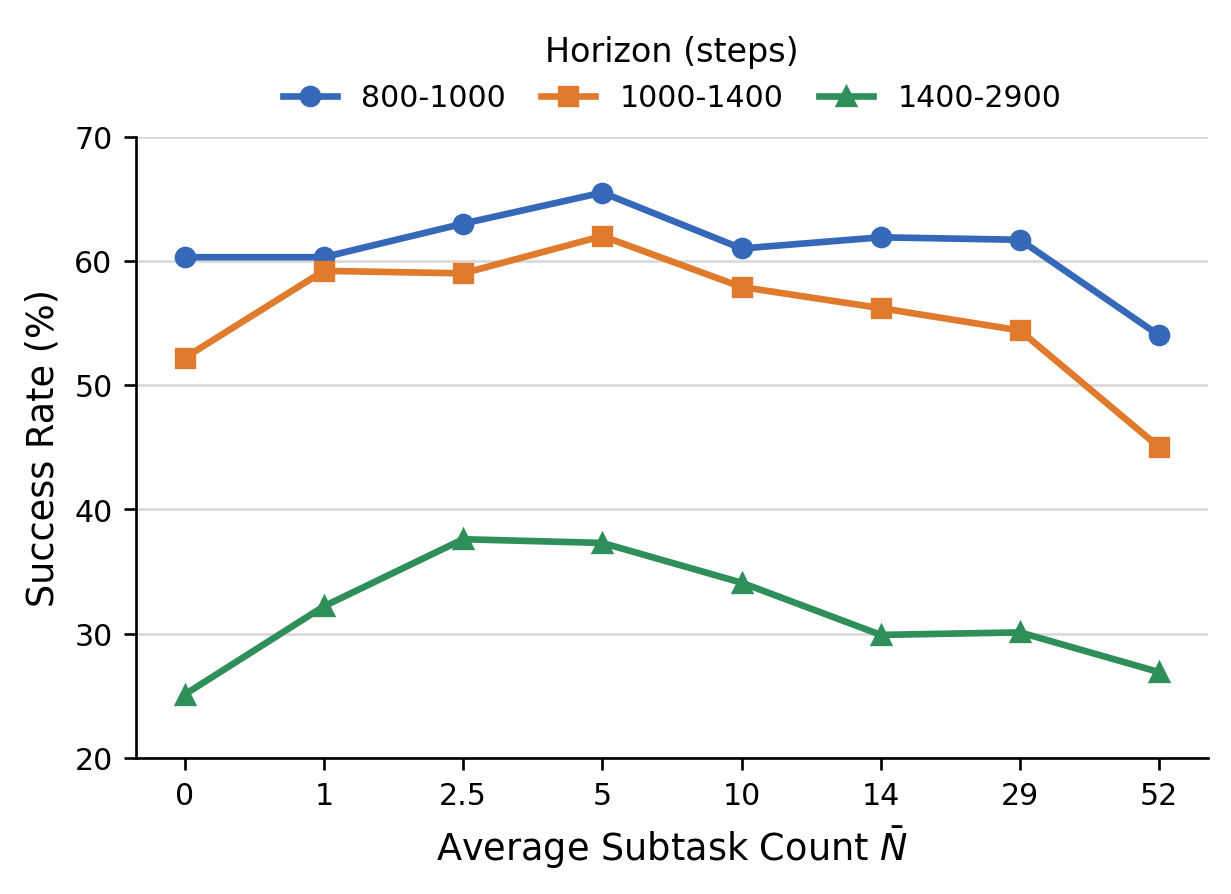}
\label{fig:density-buckets}
\end{minipage}
\hfill
\begin{minipage}[t]{0.55\textwidth}
\vspace{0pt}
\centering
\captionof{table}{SR values for different horizon buckets at each decomposition density. Each value corresponds to one point in Figure~\ref{fig:density-buckets}.}
\label{tab:density-numbers}

\setlength{\tabcolsep}{3pt}
\renewcommand{\arraystretch}{1.2}
\resizebox{\linewidth}{!}{%
\begin{tabular}{c | ccc | c}
\toprule
Avg. Subtask $\bar{N}$
 & 800--1000 & 1000--1400 & 1400--2900 & Overall\\
\midrule
0.0 & 60.3 & 52.3 & 25.1 & 40.2 \\
1.0 & 60.3 & 59.2 & 32.2 & 45.9 \\
2.375 & 63.0 & 59.0 & 37.6 & 49.1 \\
\rowcolor{tintgray} 5.0 & 65.5 & 62.0 & 37.3 & \textbf{50.4} \\
10.0 & 61.0 & 57.9 & 34.1 & 46.6 \\
14.0 & 61.9 & 56.2 & 29.9 & 44.1 \\
29.0 & 61.7 & 54.4 & 30.1 & 43.6 \\
52.0 & 54.0 & 45.0 & 26.9 & 37.6 \\
\bottomrule
\end{tabular}%
}
\end{minipage}

\end{figure}

\subsection{Effect of Decomposition Density Across Horizon Buckets}\label{app:decline_bucket}

We next investigate how decomposition density affects tasks with different horizons. Figure~\ref{fig:density-buckets} plots the SR trends for the three horizon buckets, with the exact value of each point reported in Table~\ref{tab:density-numbers}. We make two observations. First, all three horizon buckets exhibit a consistent trend: performance initially improves as the decomposition becomes denser and then declines once the density becomes too high, further confirming the overall trend observed in Figure~\ref{fig:ablation-density}. Second, comparing the best density level ($\bar{N}=5.0$) with the densest setting ($\bar{N}=52.0$), SR decreases by 11.5, 17.0, and 10.4 points for the 800--1000, 1000--1400, and 1400--2900 horizon buckets, respectively. Relative to the peak performance of each bucket, however, these correspond to declines of approximately 18\%, 27\%, and 28\%. Thus, the proportional degradation becomes larger for the two longer-horizon buckets, suggesting that excessively dense decomposition is increasingly detrimental as task horizon grows.

\subsection{Understanding Performance Degradation Under Dense Decomposition}\label{app:deccline_analysis}

Increasing subtask decomposition density changes two properties simultaneously.
First, it changes when subtask completion events are detected and rewarded.
Second, because each retained subtask can contribute an intermediate reward, it
increases the total intermediate reward available in a rollout.
Table~\ref{tab:density-mechanisms} examines these factors using controls that
hold subtask decomposition density and dependency structure fixed.

\paragraph{Signal timing and alignment.}

To isolate the effect of intermediate reward timing, we construct a controlled variant of the $\bar{N}{=}5$ decomposition while keeping the decomposition and dependency structure fixed. The reference setting uses 80 subtask-completion signals arranged into 41 ordered dependency groups. In the controlled variant, we modify only the predicate associated with each signal: the reference predicates detect the completion of each sub-motion, whereas the controlled predicates detect an earlier verifiable motion checkpoint within the same sub-motion. For example, for the straw-grasping subtask in \textit{DeliverStraw}, the reference signal fires once the grasp is complete, while the controlled signal fires at an earlier verifiable state during the same grasping motion. Thus, the subtask definition and dependencies structure remain unchanged, and only the timing at which intermediate credit is provided is shifted earlier.

We retain only earlier predicates with sufficient demonstration coverage and
non-degenerate firing behavior. This shifts 78 of the 80 completion signals earlier by an average of 3.1 action chunks; the remaining two retain their reference predicates because no valid earlier candidate exists. Under this controlled change, overall SR decreases from $50.4\%{\scriptsize\pm1.0}$ to $47.4\%{\scriptsize\pm0.8}$. This result indicates that intermediate reward timing matters even when decomposition density and dependency structure are held fixed.

\paragraph{Total intermediate reward.}
We next keep the $\bar{N}{=}14$ subtask decomposition unchanged, including all
224 subtask completion signals and their dependency structure, but cap the
total intermediate reward available in a rollout. The cap prevents total
intermediate reward from growing with the number of subtasks and raises SR from
44.1\% to 47.7\%, recovering more than half of the 6.3-point gap from the
peak of the density sweep. Thus, the performance decline at high decomposition
density is associated not only with subtask completion timing but also with the
growth of total intermediate reward.

\paragraph{Combined effect.}
The two controls isolate comparable sources of degradation. Moving the
$\bar{N}{=}5$ subtask completion signals to earlier motion checkpoints reduces
SR by 3.0 points, while removing the reward cap at $\bar{N}{=}14$ reduces SR by
approximately 3.6 points. Therefore, both subtask completion timing and total
intermediate reward should be controlled as decomposition density changes.

\begin{table}[!ht]
\centering
\caption{Controlled analyses of subtask reward timing and total intermediate reward on RoboCasa365 with GR00T-N1.5. Within each comparison, subtask decomposition density and dependency structure are fixed. Results are evaluated at training step 100 as mean $\pm$ sample standard deviation over three evaluation runs.}
\label{tab:density-mechanisms}
\setlength{\tabcolsep}{6pt}
\renewcommand{\arraystretch}{1.2}
\resizebox{0.62\textwidth}{!}{%
\begin{tabular}{l c c}
\toprule
Reward configuration & $\bar{N}$ & Overall SR (\%) \\
\midrule
\multicolumn{3}{l}{\textit{Signal timing and alignment}} \\
\quad Reference completion predicates & 5.0 & \textbf{50.4}\,{\scriptsize$\pm$1.0} \\
\quad Earlier checkpoint predicates & 5.0 & 47.4\,{\scriptsize$\pm$0.8} \\
\midrule
\multicolumn{3}{l}{\textit{Total intermediate reward}} \\
\quad Unbounded (fixed per-subtask reward) & 14.0 & 44.1\,{\scriptsize$\pm$1.8} \\
\quad Total reward capped at $B$ & 14.0 & \textbf{47.7}\,{\scriptsize$\pm$0.4} \\
\bottomrule
\end{tabular}%
}
\end{table}

\subsection{Does Reward Pacing Require Demonstration Durations?}
\label{app:demofree-pacing}

Table~\ref{tab:density-mechanisms} examined when intermediate rewards are
triggered and how their total magnitude changes with decomposition density. We
next isolate a different component: how dynamic reward pacing calibrates the
magnitude of an eligible reward after subtask completion is detected. Holding
the completion predicates, dependency structure, and decomposition density
fixed at the default decomposition (Table~\ref{tab:decomp-released}), we replace only the
demonstration-average completion duration $T_d(v_i)$ with $H/N$, where $H$ is
the task horizon and $N$ is the number of retained subtasks. This uniform
completion duration reduces SR from $49.2\%{\scriptsize\pm1.1}$ to
$43.1\%{\scriptsize\pm0.4}$ over three evaluation runs
(Table~\ref{tab:demofree-pacing}). Thus, uniform horizon allocation does not
recover the performance obtained with demonstration-derived completion
durations, while more adaptive demonstration-free estimates remain open.

\begin{table}[!ht]
\centering
\caption{Completion durations used for dynamic reward pacing at the default
subtask decomposition density. All other settings are fixed; results are mean
$\pm$ standard deviation over three evaluation runs.}
\label{tab:demofree-pacing}
\setlength{\tabcolsep}{8pt}
\renewcommand{\arraystretch}{1.15}
\resizebox{0.52\textwidth}{!}{%
\begin{tabular}{l c}
\toprule
Completion duration & Overall SR (\%) \\
\midrule
Demonstration average $T_d(v_i)$ & \textbf{49.2}\,{\scriptsize$\pm$1.1} \\
\rowcolor{tintgray}
Uniform horizon allocation $H/N$ & 43.1\,{\scriptsize$\pm$0.4} \\
\bottomrule
\end{tabular}%
}
\end{table}

\section{Additional Experimental Results}\label{app:additional}

We include additional experiments in this section, including a multi-seed evaluation (Appendix~\ref{app:random}), training dynamics (Appendix~\ref{app:RL}), and a hyperparameter sensitivity analysis (Appendix~\ref{app:hyper}).

\subsection{Multi-seed Evaluation}\label{app:random}

We first evaluate the robustness of our method across different random seeds.
Specifically, for the representative baselines in Table~\ref{tab:horizon-bucket}
and our {\name}, we conduct three independent evaluation runs with different
random seeds and report the mean and standard deviation of the resulting SR.
As shown in Table~\ref{tab:multiseed}, {\name} consistently achieves the best
overall SR on both VLA backbones. On GR00T-N1.5, {\name} reaches an overall SR
of $49.2\%$, outperforming the strongest baseline by $7.1$ points, while on
$\pi_{0.5}$ it achieves $44.8\%$, improving over the strongest baseline by
$2.9$ points. These consistent improvements across multiple random seeds
demonstrate that the gains of {\name} are robust and are not attributable to a
particular evaluation run.

\begin{table}[!ht]
\centering
\caption{Multi-seed evaluation on RoboCasa365: mean $\pm$ std over 3
independent evaluation runs, SR (\%) by horizon bucket. {\name} delivers the
best overall SR on both backbones, with the largest margins on medium and
long-horizon buckets.}
\label{tab:multiseed}
\setlength{\tabcolsep}{5pt}
\renewcommand{\arraystretch}{1.15}
\resizebox{0.66\columnwidth}{!}{%
\begin{tabular}{l | ccc | c}
\toprule
Method
 & 800--1000 & 1000--1400 & 1400--2900 & Overall\\
\midrule
\multicolumn{5}{c}{\textit{GR00T-N1.5}} \\
\quad SFT              & 52.6{\scriptsize$\pm$0.8} & 48.3{\scriptsize$\pm$0.6} & 26.6{\scriptsize$\pm$1.2} & 38.2{\scriptsize$\pm$0.5} \\
\quad w/ Sparse-RL    & 59.8{\scriptsize$\pm$1.6} & 54.3{\scriptsize$\pm$0.9} & 27.8{\scriptsize$\pm$2.9} & 42.1{\scriptsize$\pm$1.4} \\
\quad w/ SimpleVLA-RL  & 54.2{\scriptsize$\pm$2.6} & 52.6{\scriptsize$\pm$1.7} & 29.6{\scriptsize$\pm$1.2} & 41.4{\scriptsize$\pm$0.3} \\
\rowcolor{tintgray}
\quad w/ {\name}       & 63.3{\scriptsize$\pm$1.2} & 61.0{\scriptsize$\pm$1.7} & 36.5{\scriptsize$\pm$1.8} & \textbf{49.2{\scriptsize$\pm$1.1}} \\
\midrule
\multicolumn{5}{c}{\textit{$\pi_{0.5}$}} \\
\quad SFT              & 45.7{\scriptsize$\pm$5.0} & 47.9{\scriptsize$\pm$1.1} & 33.8{\scriptsize$\pm$1.2} & 40.4{\scriptsize$\pm$1.1} \\
\quad w/ Sparse-RL    & 44.2{\scriptsize$\pm$1.1} & 45.9{\scriptsize$\pm$2.2} & 35.2{\scriptsize$\pm$0.9} & 40.2{\scriptsize$\pm$0.9} \\
\quad w/ SimpleVLA-RL  & 47.8{\scriptsize$\pm$1.0} & 48.1{\scriptsize$\pm$0.8} & 35.8{\scriptsize$\pm$0.7} & 41.9{\scriptsize$\pm$0.4} \\
\rowcolor{tintgray}
\quad w/ {\name}       & 52.3{\scriptsize$\pm$2.6} & 51.3{\scriptsize$\pm$0.3} & 38.0{\scriptsize$\pm$2.2} & \textbf{44.8{\scriptsize$\pm$1.1}} \\
\bottomrule
\end{tabular}%
}
\end{table}

\subsection{Training Dynamics}\label{app:RL}

\begin{figure}[!ht]
\begin{center}
\includegraphics[width=0.95\linewidth]{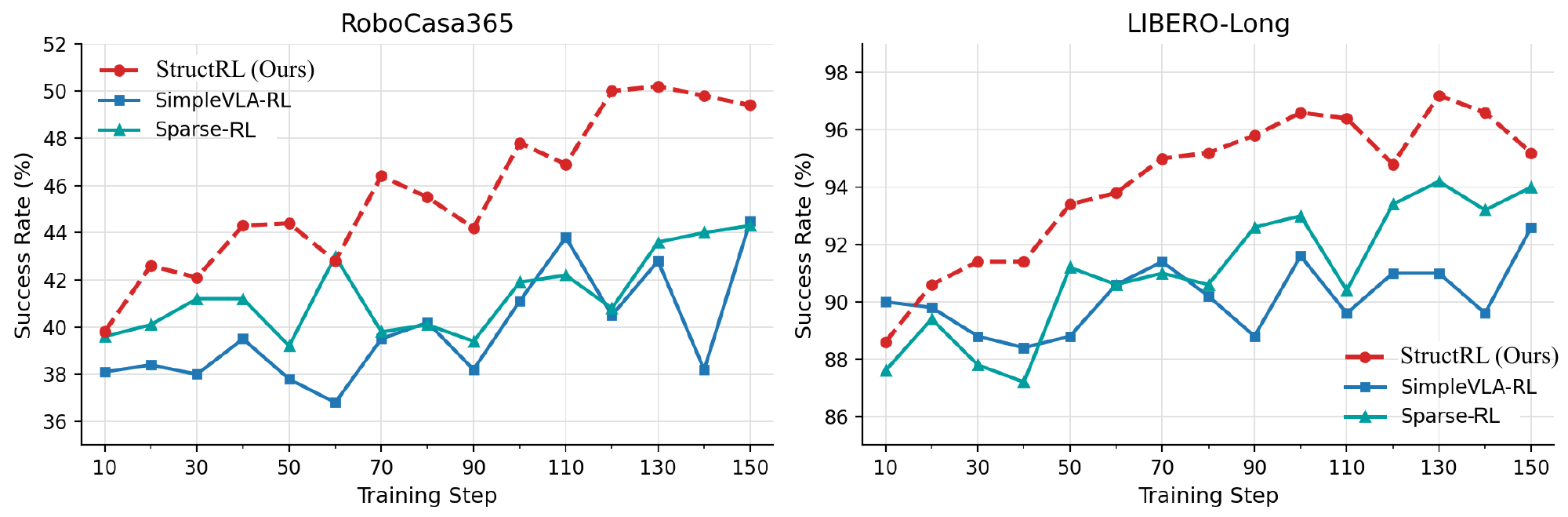}
\end{center}
\vspace{-0.1in}
\caption{\textbf{Training dynamics.} Success rate of intermediate checkpoints
(every 10 steps, offline evaluation) during RL training with GR00T-N1.5 on
RoboCasa365 composite tasks and
LIBERO-Long. {\name} consistently
outperforms SimpleVLA-RL and Sparse-RL throughout training.
}
\label{fig:training_curve}
\end{figure}

We next examine the training dynamics of {\name} to evaluate its stability throughout RL training. Specifically, we save a checkpoint every 10 training steps and evaluate each checkpoint using the corresponding benchmark evaluation protocol. We conduct this experiment with GR00T-N1.5 on both RoboCasa365 and LIBERO-Long, and continue training for up to 150 steps. The resulting training curves are shown in Figure~\ref{fig:training_curve}.

From Figure~\ref{fig:training_curve}, {\name} exhibits stable training dynamics on both benchmarks. Its performance begins to separate from the baselines after roughly 50 training steps and remains consistently higher throughout the remainder of training, without any performance collapse as training proceeds. Moreover, the performance gains are already evident by around 100 training steps and persist afterward. Therefore, considering the additional computational cost of longer training, we uniformly report the results at training step 100 in our main experiments.

\begin{figure}[!ht]
\centering
\vspace{-4pt}
\includegraphics[width=0.62\linewidth]{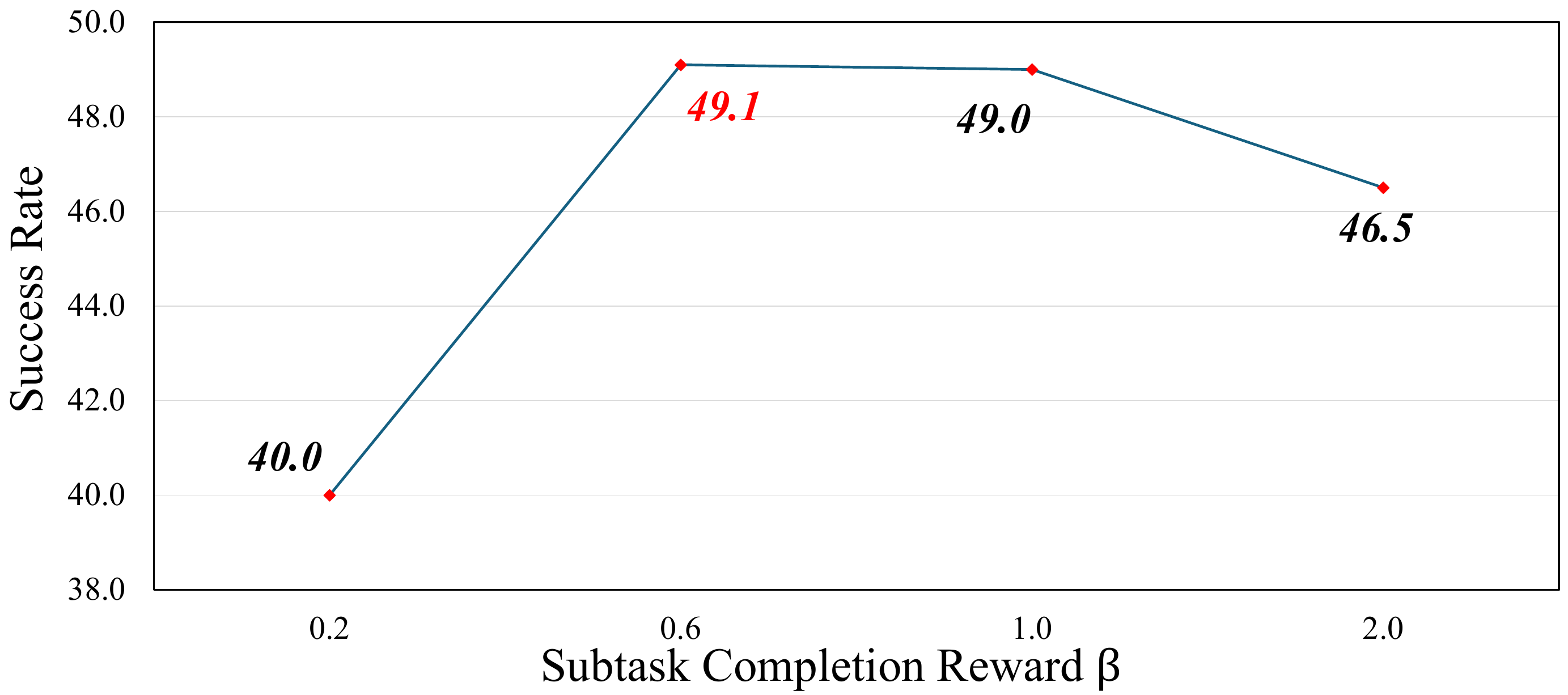}
\vspace{-4pt}
\caption{
Sensitivity of {\name} to the subtask-completion reward $\beta$ on
RoboCasa365 using GR00T-N1.5. The terminal reward is held fixed.
}
\label{fig:beta-mass}
\vspace{-6pt}
\end{figure}

\subsection{Sensitivity to the Subtask-completion Reward}
\label{app:hyper}

In Eq.~\eqref{eq:pace}, we use the hyperparameter $\beta$ to control the magnitude of the dynamically paced subtask reward. We therefore examine the sensitivity of {\name} to the choice of $\beta$. Specifically, we fix the terminal reward $\lambda_c$ to $2.0$ and vary $\beta$ over $\{0.2, 0.6, 1.0, 2.0\}$. We conduct the experiments on RoboCasa365 using GR00T-N1.5 and report the resulting SR in Figure~\ref{fig:beta-mass}. As shown in the figure, performance improves substantially when $\beta$ increases from $0.2$ to $0.6$, with SR increasing from $40.0\%$ to $49.1\%$. Performance remains nearly unchanged at $\beta=1.0$, reaching $49.0\%$, while a larger value of $\beta=2.0$ reduces SR to $46.5\%$. These results indicate that {\name} is relatively insensitive to $\beta$ within a moderate range, while assigning either too little or too much weight to subtask rewards can degrade performance.

\section{Case Study}\label{app:case_study}

\begin{figure}[!ht]
\centering
\includegraphics[width=0.95\linewidth]{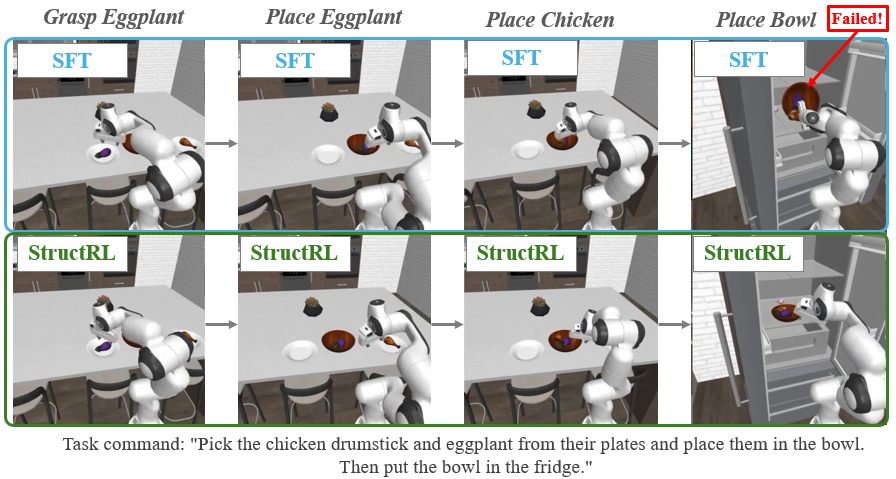}
\caption{Qualitative comparison between SFT and {\name} on the long-horizon task ``StoreLeftoversInBowl''. {\name} successfully completes the intermediate object-placement steps and proceeds to placing the bowl in the refrigerator, while SFT fails before reaching the final stage.}
\label{fig:case1}
\end{figure}

\begin{figure}[!ht]
\centering
\includegraphics[width=0.95\linewidth]{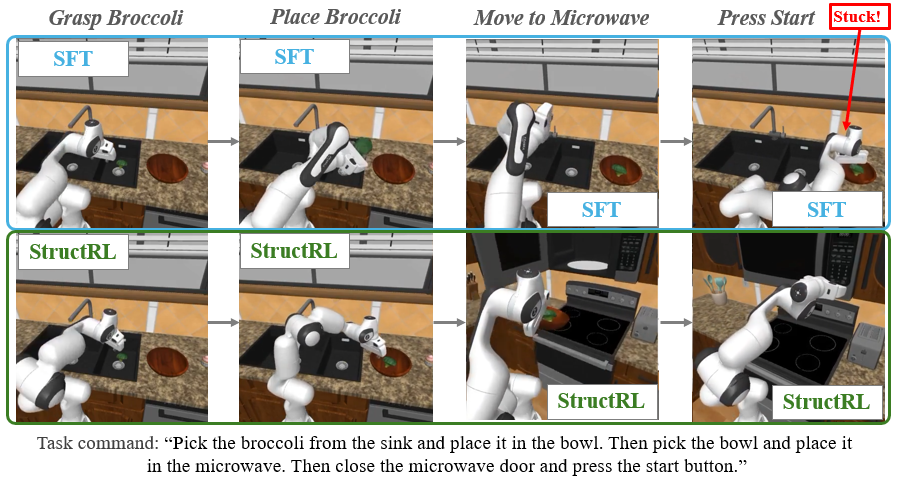}
\caption{Qualitative comparison between SFT and {\name} on the long-horizon task ``SteamInMicrowave''. {\name} completes the preceding subtasks and advances to interacting with the microwave, whereas SFT gets stuck at an earlier manipulation stage.}
\label{fig:case2}
\end{figure}

We further provide qualitative case studies to compare the execution behaviors
of the SFT policy and {\name} on long-horizon VLA tasks. As shown in
Figures~\ref{fig:case1} and~\ref{fig:case2}, each example presents the task
command together with a sequence of observations from the SFT policy and
{\name} at several key moments during execution. By tracking the same agent
view over time, the examples make it easier to compare how the two policies
progress through the task and where their behaviors begin to diverge.

In Figure~\ref{fig:case1}, both policies begin by manipulating the target
objects, but {\name} continues to complete the required intermediate subtasks
and eventually reaches the final stage of placing the bowl into the
refrigerator, whereas the SFT policy fails to make comparable progress. In
Figure~\ref{fig:case2}, {\name} successfully progresses from grasping and
placing the broccoli to moving the bowl into the microwave, while the SFT
policy becomes stuck at an earlier manipulation stage. These qualitative
examples illustrate that {\name} makes more consistent progress through
long-horizon tasks and is less likely to get stuck at intermediate stages than
the SFT policy.

\section{Robometer Baseline Implementation}
\label{app:robometer}

We implement the learned dense-reward baseline using the released Robometer-4B checkpoint~\citep{liang2026robometer}, without additional fine-tuning. Its training corpus includes LIBERO-Long~\citep{liu2023libero} demonstrations and failure trajectories generated on the same tasks~\citep{liang2026robometer}, so on LIBERO-Long the reward model is evaluated in-distribution. At each action-chunk boundary $k$, Robometer receives the task instruction and a causal visual history and predicts scalar progress $p_k\in[0,1]$, computed as the expectation over its ten discrete progress bins. We convert this prediction to a chunk-level reward as
\begin{equation}
r_k^{\mathrm{Robo}}
=\beta_{\mathrm{R}}\bigl(p_k-b_k\bigr)
+\lambda_c\,\mathbbm{1}[s_k=1],
\qquad
b_k=
\begin{cases}
0, & s_k=1,\\
1, & s_k=0,
\end{cases}
\label{eq:robometer-reward}
\end{equation}
where $s_k=1$ only when task success is first detected during chunk $k$, and $\lambda_c=2.0$ is the same terminal reward used by {\name}. Before success, the reward is therefore $\beta_{\mathrm{R}}(p_k-1)$; at the first successful chunk, it becomes $\beta_{\mathrm{R}}p_k+\lambda_c$. This follows the online-RL reward formulation used with Robometer~\citep{liang2026robometer}, adapted from simulator-step to action-chunk granularity. The reward is assigned once at each chunk boundary, with zero reward assigned to the remaining simulator steps within that chunk. No further reward is emitted after the first detected success. We use the raw progress estimates without temporal differencing or clipping.

For each query, the current boundary frame is appended to a causal history maintained separately for each environment. Robometer receives $T=8$ frames, matching its training-time input length, sampled at approximately uniform indices $\lfloor\operatorname{linspace}(0,t,8)\rceil$ from the first frame through the current boundary. This construction uses no future observations. Robometer is queried once per action chunk, corresponding to every 16 simulator steps for GR00T-N1.5 and every 10 steps for $\pi_{0.5}$.

On LIBERO-Long, we set $\beta_{\mathrm{R}}=0.07041$ for GR00T-N1.5 and $\beta_{\mathrm{R}}=0.04401$ for $\pi_{0.5}$. These coefficients are selected once before RL training using reference rollouts to match the intermediate-reward scale of {\name}. All other PPO hyperparameters, interaction budgets, terminal rewards, and evaluation settings are shared with {\name}. Applying the negative base once per chunk prevents unsuccessful policies from accumulating positive progress reward merely by extending a rollout, while stopping reward emission after success prevents repeated credit from a latched success state.

\section{Limitations}

In this work, we introduced {\name}, a structured online RL framework that improves long-horizon VLA training by combining verifiable subtask supervision with structure-aware reward gating and dynamic reward pacing. While our experiments demonstrate strong empirical performance, several directions remain for further study.

\noindent\textbf{Subtask completion signals.}
Our main experiments use simulator predicates to provide precise and reproducible detection of subtask completion. In real-world settings, these signals would need to be derived from available sensory observations or other task-specific feedback. Extending {\name} to such settings is an important direction for future work and may involve integrating suitable progress-detection mechanisms.

\noindent\textbf{Real-world evaluation.}
Our evaluation focuses on simulation, which enables controlled and reproducible comparisons across diverse long-horizon tasks and training configurations. Evaluating {\name} on physical robotic systems is a natural next step to study its behavior under real-world perception, dynamics, and execution variability.

\end{document}